\documentclass[10pt,twocolumn,letterpaper]{article}

\usepackage{wacv}                       % camera-ready (do NOT use for review)

\usepackage{amsfonts}
\usepackage{multirow}
\usepackage{placeins}   % \FloatBarrier — keep each results figure in its own subsection
\graphicspath{{figures/}}
\definecolor{figgreen}{HTML}{1A9850}
\definecolor{figred}{HTML}{D73027}
\definecolor{figblue}{HTML}{4575B4}


\definecolor{wacvblue}{rgb}{0.21,0.49,0.74}
\usepackage[pagebackref,breaklinks,colorlinks,allcolors=wacvblue]{hyperref}

\def\wacvPaperID{2251} % enter the CMT Paper ID here before submission
\def\confName{WACV}
\def\confYear{2027}

\title{A Decision-Aware Evaluation Protocol for Uncertainty \\ in Medical Image Segmentation}

\author{Saket Maganti$^1$ \quad Shreyank N Gowda$^2$ \\
{\footnotesize $^1$Independent Researcher} \\  {\footnotesize $^2$School of Computer Science, University of Nottingham, Nottingham, UK}\\
{\tt\footnotesize saket.20bce7367@vitapalum.ac.in \quad shreyank.narayanagowda@nottingham.ac.uk}}

\begin{document}
\maketitle

\begin{abstract}
Medical segmentation uncertainty is often evaluated with calibration error or pixel-level error-detection scores, but these metrics do not show whether uncertainty improves downstream review decisions. We propose a decision-aware evaluation protocol that separates uncertainty estimation from deferral policy and evaluates their combination under fixed review budgets: which pixels, regions, or images should be accepted, and which should be reviewed. The protocol reports fixed-budget error reduction, risk-coverage behaviour, region-level review proxies, structural failure modes, probability-derived baselines, and sensitivity to review assumptions. We instantiate it across retinal vessel, skin lesion, and polyp segmentation with multiple backbones and uncertainty estimators. Under matched inference budgets and canonical multi-layer MC Dropout, test-time augmentation (TTA) gives a stronger error-detection signal across matched MC--TTA comparisons, including architecture and dropout-placement checks. However, estimator rankings are not universal. For example, probability-derived scores remain competitive, deep ensembles can outperform TTA, learned deferral helps mainly when the base estimator is weak, and rankings can change with review unit, uncertainty target, correction assumptions, and cost model. Structural and second-observer analyses further show that boundary and thin-structure failures are partly tied to annotation ambiguity. These findings support evaluating uncertainty as an estimator-policy-review configuration, not as a standalone score map. Code: \url{https://anonymous.4open.science/r/Uncertainity-in-Medical-Image-Segmentation/README.md}
\end{abstract}

\section{Introduction}
\label{sec:intro}

A medical segmentation model is rarely the final actor in a clinical or
annotation workflow. Its mask is inspected, corrected, accepted, or escalated.
What matters is therefore not only whether the model is accurate on average, but
whether it can indicate when its output should not be trusted. This distinction
is easy to miss. A model may report a high Dice score while failing on the cases
that make review necessary: ambiguous boundaries, thin vessels, small polyps, or
images with artefacts. In such settings, uncertainty is valuable only if it
changes the downstream decision of which pixels, regions, or images should be
accepted and which should be reviewed under a limited budget.

Most work on uncertainty in medical segmentation evaluates the uncertainty map as
an object in itself. Calibration error, predictive likelihood, and pixel-level
error-detection AUROC
\cite{gal2016dropout,lakshminarayanan2017simple,mehrtash2020confidence}
ask whether uncertainty values are numerically meaningful or whether they rank
mistakes well. These are important questions, but they are not the whole
deployment question. Two uncertainty scores with similar AUROC can select
different pixels at a 10\% review budget. A better calibrated model can still
leave more residual error after deferral. A segmentation model without an
explicit review rule has no operating point: it has a score map, but not a
decision.

This paper argues that segmentation uncertainty should be evaluated at the point
where it is used. We separate \textbf{estimation}, the procedure that produces an
uncertainty or confidence field, from \textbf{decision}, the policy that converts
that field into accept/defer choices. This separation turns uncertainty
evaluation into a practical question: under the same review budget, which
estimator and policy remove more error, retain safer predictions, and expose the
failure modes that would otherwise remain hidden? It also changes what counts as
a fair baseline. In binary segmentation, common probability-derived scores such
as entropy, margin, inverse confidence, and distance from $1/2$ collapse to the
same ranking. These free scores should therefore be treated as first-class
comparators, not as weak baselines.

We propose a decision-aware evaluation protocol for medical image segmentation.
The protocol pairs uncertainty estimators with deferral policies and reports
fixed-budget error reduction, risk--coverage behaviour, region-level review
proxies, structural stratification, probability-derived baselines, and
sensitivity to review assumptions under the same decision rule. It is an
evaluation practice rather than a new architecture or reject rule: any backbone,
uncertainty estimator, or review policy can be inserted, but all are judged by
the decisions they enable under explicit review budgets.

We instantiate the protocol across retinal vessel, skin lesion, and polyp
segmentation with multiple backbones and uncertainty estimators. The results show
that estimator rankings are decision-dependent. Under matched inference budgets
and canonical multi-layer MC Dropout, test-time augmentation (TTA) gives a
stronger error-detection signal across matched MC--TTA comparisons, including
complementary-architecture and dropout-placement checks. However, TTA is not a
universal winner: probability-derived scores remain competitive, deep ensembles
can outperform TTA on some datasets, and learned deferral heads help mainly when
the base estimator is weak. Review assumptions also matter: rankings can change
with the review unit, uncertainty target, correction model, and cost function.
Structural and second-observer analyses further show that boundary and
thin-structure failures are partly tied to annotation ambiguity, but are not
explained away by it.

\paragraph{Contributions.}
\begin{enumerate}[leftmargin=*,itemsep=2pt,topsep=2pt]
  \item We introduce a \textbf{decision-aware evaluation protocol} for medical
  image segmentation uncertainty, separating uncertainty estimation from review
  policy and evaluating their combination through fixed-budget error reduction,
  risk--coverage AURC, region-unit proxies, structural stratification, and
  review-assumption sensitivity.

  \item We provide a \textbf{controlled empirical benchmark} across multiple
  datasets, backbones, uncertainty estimators, and policies, including
  probability-derived scores as first-class baselines and matched-budget
  comparisons of canonical MC Dropout and TTA.

  \item We show that \textbf{uncertainty rankings are not universal}: TTA is
  robustly stronger than canonical MC Dropout under matched budgets, but deep
  ensembles, probability baselines, learned policies, review units, uncertainty
  targets, and annotation assumptions can change the preferred decision rule.
\end{enumerate}

\section{Related Work}
\label{sec:related_work}

\textbf{Medical segmentation models and foundation models.}
Medical segmentation has progressed from U-Net-style CNNs
\cite{ronneberger2015unet,zhou2018unetplusplus} to self-configuring pipelines
\cite{isensee2021nnunet}, transformer hybrids \cite{chen2021transunet}, vision-language foundation models \cite{liu2024g2d,gowda2024masks,medklip} and
promptable foundation models
\cite{kirillov2023segment,ma2024medsam,ravi2025sam2,cc-sam}, whose zero-shot medical
use has been benchmarked across modalities \cite{mazurowski2023segment}. These
works ask how to build or adapt stronger segmenters. Our question is
orthogonal: given a trained segmenter, how should its uncertainty be evaluated
when it drives review decisions? We therefore hold the backbone fixed within
each dataset--backbone cell when comparing estimators and policies, and include
complementary architecture checks to test whether the observed decision patterns
are tied to one backbone family. A SAM/MedSAM extension treats the prompt as an
additional decision variable, but keeps promptable models outside the controlled
trained-backbone matrix.

\textbf{Uncertainty estimation for segmentation.}
Uncertainty is commonly estimated with MC Dropout \cite{gal2016dropout}, deep
ensembles \cite{lakshminarayanan2017simple}, evidential methods
\cite{sensoy2018evidential,ulmer2023prior}, or test-time perturbations such as
TTA \cite{ayhan2018test,wang2019aleatoric,moshkov2020tta}, where aggregation can
change the estimate \cite{shanmugam2021better}. In medical segmentation, prior
work has studied confidence calibration \cite{mehrtash2020confidence}, whether
uncertainty helps downstream tasks
\cite{czolbe2021segmentation,band2021benchmarking,nair2020exploring}, and
robustness under distribution shift \cite{ovadia2019trust}. We treat these
estimators as inputs to a review decision rather than as endpoints: MC Dropout,
TTA, probability-derived scores, and deep ensembles are compared by the
decisions they enable under matched budgets and review assumptions.

\textbf{Calibration, confidence, and decision metrics.}
Calibration \cite{guo2017calibration} measures agreement between confidence and
accuracy, while selective prediction \cite{geifman2017selective} measures the
risk retained after abstention. A better calibrated model need not produce a
better review set at a fixed budget. Related work predicts errors directly, from
maximum-softmax confidence \cite{hendrycks2017baseline} to learned confidence
heads \cite{corbiere2019addressing} and reliability analyses for medical
segmentation \cite{jungo2019assessing}. Decision curve analysis
\cite{vickers2006decision,vickers2019dca} similarly evaluates the benefit of
acting on predictions. Closest in motivation, \citet{czolbe2021segmentation}
ask whether segmentation uncertainty helps downstream tasks, and
\citet{jungo2019assessing} audit uncertainty failures. We differ by making the
estimator--policy pair the evaluation unit and by requiring probability-derived
baselines, fixed-budget error reduction, risk--coverage, structural
stratification, region-unit behaviour, and review-assumption sensitivity across
a multi-domain benchmark.

\textbf{Abstention, deferral, and conformal guarantees.}
Selective prediction and learning to defer formalise abstention or referral
under review cost
\cite{chow1970optimum,geifman2019selectivenet,mozannar2020consistent,
verma2022calibrated,mozannar2023who}. Conformal methods provide finite-sample
guarantees \cite{angelopoulos2021gentle,angelopoulos2024conformal}, including
risk-controlling prediction sets \cite{bates2021rcps} and dense-prediction
extensions \cite{mossina2024conformal}. These methods provide possible reject
rules, guarantees, or operating points. Our contribution is complementary: we do
not propose a new reject-rule theory or a new uncertainty estimator. Instead, we
provide a benchmark protocol that assembles existing estimators, baselines,
review units, and deferral policies into a common decision-aware evaluation
framework.

\begin{figure*}[t]
  \centering
  \includegraphics[width=\textwidth]{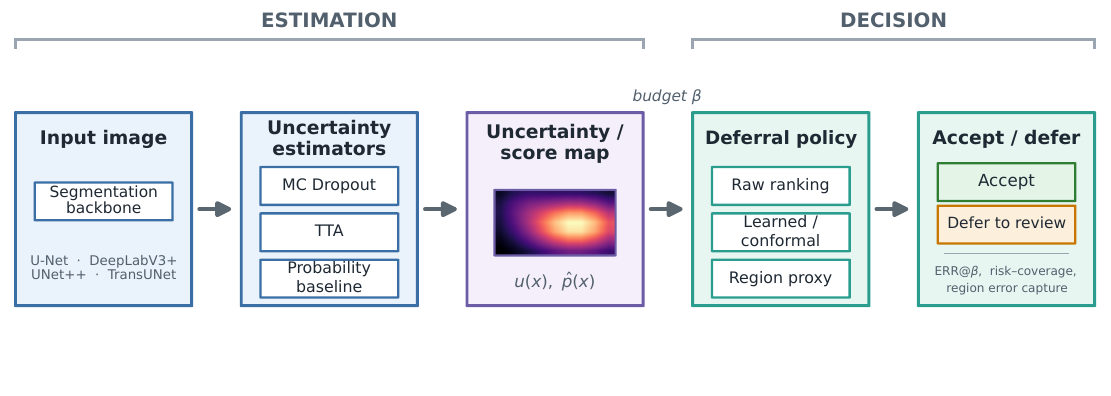}
  \caption{Decision-aware evaluation separates uncertainty \emph{estimation}
  from the review \emph{decision}. A trained segmentation backbone produces
  $\hat{p}$ and $\hat{y}$; an estimator, ensemble, or probability-derived
  baseline produces a score map $u$; and a deferral policy ranks pixels or
  regions under budget $\beta$. ERR@$\beta$, risk--coverage AURC, region error
  capture, and review-assumption sensitivity are measured after this decision,
  not from the uncertainty map alone.}
  \label{fig:pipeline}
\end{figure*}

\section{Decision-Aware Evaluation Protocol}
\label{sec:method}

We evaluate segmentation uncertainty as part of a review decision, not as a
standalone map. The protocol has two stages (\cref{fig:pipeline}). First, an
\emph{estimator} takes an image and a trained segmentation backbone and returns
a probability map $\hat{p}$, a hard prediction $\hat{y}$, and an uncertainty or
score map $u$. Second, a \emph{deferral policy} ranks review units such as
pixels, patches, connected components, or images, and at budget $\beta$ accepts
the $1-\beta$ fraction with the lowest review score while deferring the rest.
Thus, the protocol separates how uncertainty is estimated from how it is used.

\subsection{Uncertainty estimators and score maps}
\label{sec:estimators}

Within each dataset--backbone cell, the segmentation backbone is held fixed when
comparing estimators and policies. \textbf{MC Dropout}
\cite{gal2016dropout} keeps dropout active at test time, freezes normalization
layers, averages stochastic forward passes, and reports predictive variance or
mutual information. Our primary MC--TTA comparison uses canonical multi-layer MC
Dropout under matched inference budgets; dropout placement, dropout rate,
stochastic realization, and training seed are audited separately. \textbf{TTA}
\cite{ayhan2018test} averages label-preserving geometric transforms after
inverse warping and uses variation across transformed predictions as
uncertainty. We evaluate matched TTA subsets and the full six-transform set.

We also include \textbf{deep ensembles}
\cite{lakshminarayanan2017simple}. Because an ensemble combines independently
trained models, it is reported as a stronger estimator-family comparison rather
than as a within-checkpoint perturbation. Finally, we include probability-derived
scores as first-class baselines: entropy, inverse confidence, margin, distance
from $1/2$, and maximum-softmax confidence. For binary segmentation, these
scores are monotone functions of $|\hat{p}-1/2|$ and induce the same ranking, so
we report a single probability-baseline curve.

\subsection{Deferral as error-probability ranking}
\label{sec:risk}

Let $x$ be a review unit and
$\eta(x)=P(\hat{y}(x)\neq y(x)\mid x)$ its probability of error. A policy
$\delta(x)\in\{0,1\}$ accepts ($1$) or defers ($0$). For error loss $\ell$ and
review cost $\lambda$, the selective risk is
\begin{equation}
  R(\delta;\lambda)
  =
  \mathbb{E}\!\left[\ell\,\delta(x)\right]
  +
  \lambda\,\mathbb{E}\!\left[1-\delta(x)\right],
  \label{eq:selective_risk}
\end{equation}
that is, error retained among accepted units plus the cost of review. The
pointwise optimum is the Chow reject rule \cite{chow1970optimum},
\[
  \delta^\star(x)=1 \Longleftrightarrow \eta(x)\leq \lambda .
\]
Thus, every deferral policy is implicitly ranking units by their probability of
being wrong. Our contribution is not a new reject rule, but a protocol for
evaluating estimators and policies under explicit review budgets.

\subsection{Deferral policies}
\label{sec:policies}

We evaluate policies under the same fixed-budget accept/defer rule.
\textbf{Raw uncertainty} ranks units by $u$. \textbf{Confidence-aware
uncertainty} ranks by $u(1-c)$, where
$c=2|\hat{p}-\tfrac{1}{2}|$ is binary confidence. \textbf{Probability baselines}
rank by the softmax-derived score above. \textbf{Adaptive thresholding} selects
operating points on validation data while preserving the fixed review budget.
\textbf{Learned deferral heads} estimate $\eta$ from validation errors using
uncertainty, confidence, entropy, local statistics, and gradient features while
keeping the segmentation backbone frozen; test errors are never used to fit the
policy. We report validation-set sizes because small validation sets can make
learned policies data-starved. \textbf{Conformal thresholds} are supported as a
protocol interface for split-conformal or conformal-risk-control operating
points \cite{angelopoulos2024conformal}, although our comparisons focus on raw,
probability-derived, adaptive, learned, and ensemble-based policies.

\subsection{Decision metrics and review assumptions}
\label{sec:metrics}

All metrics are computed after applying the accept/defer decision.
\textbf{Unc-AUROC} is the score-level bridge to prior work: pixel-level ROC AUC
for detecting segmentation errors. \textbf{ERR}$(\beta)$ measures relative error
reduction among accepted units at budget $\beta$, reported with accepted Dice and
the fraction of all errors captured by review. \textbf{Risk--coverage AURC}
integrates accepted error over coverage, where lower is better.

Fixed-budget error capture assumes deferred units can be corrected, so it is an
analytical review proxy rather than a reader-study measurement. We therefore
audit sensitivity to reviewer quality, review unit, uncertainty target,
threshold, morphology definition, and cost model. We also report structural
stratification for boundary, interior, and thin-structure pixels, and use
region-unit proxies such as $16\times16$ blocks, top-$k$ patches, and connected
components to test whether uncertainty remains useful when review is organised
over regions rather than isolated pixels.

\section{Experimental Setup}
\label{sec:experiments}

\subsection{Datasets, backbones, and splits}
\label{sec:datasets}

We instantiate the protocol across retinal vessel, dermoscopic lesion, and
colonoscopy polyp segmentation using DRIVE \cite{staal2004drive}, FIVES
\cite{jin2022fives}, ISIC~2018 \cite{codella2019isic}, and Kvasir-SEG
\cite{jha2020kvasir}. DRIVE uses the official 20-image test set with the
first-observer annotation; its train/validation split is self-defined. FIVES,
ISIC, and Kvasir use self-defined 65/15/20 splits with a fixed seed. We report
all train/validation/test counts, split hashes, and test identifiers in the
supplement. Summary in Tab.~\ref{tab:splits_main}.

The primary matrix spans U-Net \cite{ronneberger2015unet}, UNet++
\cite{zhou2018unetplusplus}, DeepLabV3+ \cite{chen2018deeplabv3plus}, and
TransUNet \cite{chen2021transunet}. Within each dataset--backbone cell, the
segmentation checkpoint is fixed when comparing estimators and policies.
Additional complementary-architecture checks test whether the observed rankings
are tied to one backbone family. 

\begin{table}[t]
\centering
\small
\setlength{\tabcolsep}{4pt}
\caption{\textbf{Dataset splits and fitting data.} DRIVE uses the official
20-image test set with a self-defined train/validation split. FIVES, ISIC, and
Kvasir use fixed 65/15/20 splits. Learned heads and adaptive policies are fit
only on validation images.}
\label{tab:splits_main}
\resizebox{\columnwidth}{!}{%
\begin{tabular}{lcccc}
\toprule
Dataset & Train & Val / fit & Test & Split status \\
\midrule
DRIVE & 16 & 4 & 20 & official test \\
FIVES & 520 & 120 & 160 & self-defined \\
ISIC 2018 & 1686 & 389 & 519 & self-defined \\
Kvasir-SEG & 650 & 150 & 200 & self-defined \\
\bottomrule
\end{tabular}%
}
\end{table}

\subsection{Uncertainty inference and policies}
\label{sec:training_inference}

The main MC--TTA comparison uses canonical multi-layer MC Dropout with frozen
normalization layers and matched inference budgets. MC Dropout reports
predictive variance or mutual information from stochastic passes, while TTA
reports variation across label-preserving geometric transforms after inverse
warping. We also audit stochastic repetitions, TTA subsets, dropout placement,
dropout rate, training seed, and five-member deep ensembles.

Deferral policies are selected or fitted only on held-out validation data from
the corresponding cell. Learned deferral heads estimate error probability from
validation errors with the segmentation network frozen; test labels are never
used to fit a policy. Probability-derived baselines and raw uncertainty policies
require no additional training. Validation sizes are reported explicitly because
small validation sets, especially DRIVE, can make learned policies data-starved.

\subsection{Statistical analysis and provenance}
\label{sec:statistics}

All decision metrics are computed on paired held-out images. Where image-level
paired data are available, we use image-level bootstrap confidence intervals,
paired sign-flip tests, exact sign tests, and Benjamini--Hochberg correction
within named comparison families. Cross-cell counts are reported as descriptive
summaries rather than as independent population samples.
All reported comparisons are backed by traceable artifacts including aligned image IDs,
per-image metrics, split metadata, and table-generation records. The supplement
provides the full artifact audit, numerical tables, negative and blocked results,
and checks for learned deferral, policy transfer, second-observer analysis, and
review-assumption sensitivity.
We distinguish descriptive cross-cell counts from paired image-level inference:
counts summarize the evaluated benchmark, while uncertainty estimates and
significance tests are computed on held-out images within each declared
comparison family.

\section{Results}
\label{sec:results}

We report results as decision-level findings. Under matched-budget controls,
TTA gives a stronger error-detection signal than canonical MC Dropout, but the
broader evaluation shows that estimator rankings are not universal. Rankings
change with estimator family, review unit, uncertainty target, correction
assumption, and cost model. The main result is therefore not a single winning
uncertainty method, but the need to evaluate uncertainty as an
estimator--policy--review configuration.

The results are organized accordingly: we first establish the controlled
matched-budget comparison, then show where that comparison is stable, where it
breaks under different estimator families or budgets, and where decision-layer
choices change the conclusion even with predictions fixed.

\subsection{Matched-budget controls strengthen the MC--TTA comparison}
\label{sec:results_mctta}

\begin{table}[t]
\centering
\small
\setlength{\tabcolsep}{4pt}
\caption{\textbf{Matched-budget MC--TTA robustness.} Each row is a paired
image-level comparison family with BH correction applied within that family.
Positive cells favour TTA.}
\label{tab:mctta_robustness}
\resizebox{\columnwidth}{!}{%
\begin{tabular}{lccc}
\toprule
Comparison family & Positive & BH sig. & Mean $\Delta$ \\
\midrule
Matched budget, anchor & 24/24 & 24/24 & +0.035 \\
10 MC realizations & 24/24 & 24/24 & +0.033 \\
Complementary backbone & 24/24 & 24/24 & +0.063 \\
Dropout placement & 72/72 & 72/72 & +0.075 \\
Dropout rate, DRIVE only & 24/24 & 18/24 & +0.015 \\
Training seed, DRIVE only & 15/18 & 10/18 & +0.009 \\
\bottomrule
\end{tabular}%
}
\end{table}

This direction persists under the main implementation controls
(\cref{tab:mctta_robustness}). Ten independent MC stochastic realizations,
all TTA transform subsets, complementary-architecture checks, and
dropout-placement experiments preserve the TTA advantage. The controls also
narrow the claim: head-only dropout materially underestimates MC quality, while
DRIVE-only dropout-rate and seed sweeps show smaller effects and low-budget
ties or reversals. Unequal-budget stress tests further show that giving MC many
more stochastic samples can produce ties or reversals. We therefore state the
result as a matched-budget, canonical-dropout finding, not as a
compute-independent ordering.

\subsection{No estimator is universally dominant}
\label{sec:results_estimators}

\begin{table}[t]
\centering
\small
\setlength{\tabcolsep}{4pt}
\caption{\textbf{Estimator family changes the ranking.} Mean image-level
error-detection AUROC on the same held-out images. TTA uses six transforms.
$\Delta$ values are paired differences against TTA-T6; positive means the
competitor is better. Asterisks denote BH-corrected significance within the
comparison family.}
\label{tab:estimator_family}
\resizebox{\columnwidth}{!}{%
\begin{tabular}{llccccc}
\toprule
Dataset & Func. & MC-T6 & MC-T50 & TTA-T6 & Ens. & $\Delta_{\mathrm{Ens}}$ \\
\midrule
DRIVE  & Var & 0.839 & 0.857 & 0.858 & 0.806 & $-0.052^{*}$ \\
DRIVE  & MI  & 0.852 & 0.868 & 0.868 & 0.799 & $-0.069^{*}$ \\
FIVES  & Var & 0.852 & 0.878 & 0.858 & 0.881 & $+0.023^{*}$ \\
FIVES  & MI  & 0.859 & 0.882 & 0.872 & 0.886 & $+0.014^{*}$ \\
ISIC   & Var & 0.803 & 0.840 & 0.867 & 0.896 & $+0.029^{*}$ \\
ISIC   & MI  & 0.821 & 0.851 & 0.874 & 0.894 & $+0.019^{*}$ \\
Kvasir & Var & 0.832 & 0.866 & 0.897 & 0.908 & $+0.011$ \\
Kvasir & MI  & 0.864 & 0.889 & 0.911 & 0.907 & $-0.004$ \\
\bottomrule
\end{tabular}%
}
\end{table}

Deep ensembles provide a stronger estimator-family comparison. Because an
ensemble combines independently trained models, it changes the segmenter family
rather than perturbing a fixed checkpoint. As shown in
\cref{tab:estimator_family}, the result is not a simple ``TTA-best'' ordering:
TTA is significantly better than the ensemble on DRIVE, the ensemble is
significantly better on FIVES and ISIC, and Kvasir is not significantly
different after correction. This supports the protocol's main premise:
estimator rankings are empirical and dataset-dependent.

\begin{figure*}[t]
  \centering
  \includegraphics[width=\textwidth]{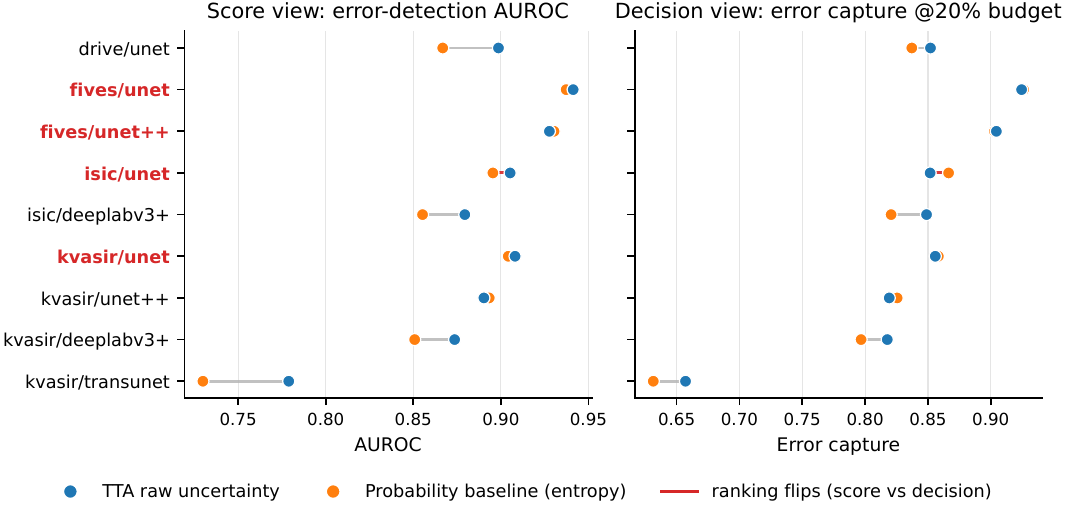}
  \caption{\textbf{The score ranking is not the decision ranking.} Per cell, TTA
  raw uncertainty and the probability baseline are compared under a score-level
  metric (left: error-detection AUROC) and a decision-level metric (right:
  fraction of all errors captured at a $20\%$ review budget). Rankings can flip
  at the decision point.}
  \label{fig:caldec}
\end{figure*}

\subsection{Score-level rankings do not determine review decisions}
\label{sec:score_vs_decision}

Probability-derived baselines remain important even after the matched-budget
MC--TTA controls. In binary segmentation, entropy, margin, inverse confidence,
maximum-softmax confidence, and distance from $1/2$ collapse to one ranking,
which is available without extra inference. In the nine-cell decision matrix,
raw TTA uncertainty exceeds this probability baseline in seven cells by a small
mean margin, while the baseline wins in two.

More importantly, the ranking can change at the decision point
(\cref{fig:caldec}). AUROC and 20\%-budget error capture disagree in four of
nine cells, and AUROC-margin versus capture-margin is only moderately
rank-correlated (Spearman $\rho=0.70$, Kendall $\tau=0.50$). Calibration is an
even weaker proxy: ECE is not positively correlated with error-detection AUROC
across cells. Thus, calibration, error ranking, and review utility are related
but distinct evaluation targets.

\subsection{Learned deferral helps, but is not a free win}
\label{sec:results_fixed_budget}

Learned deferral heads estimate error probability from validation errors while
keeping the segmentation backbone frozen. Under the deployable validation-fit
protocol, the learned head improves Capt.@20\% in 16/18 cell--arm combinations,
is significantly worse than plain uncertainty in one, and is not significant in
one. The gains are largest for weaker MC-based rankings; for example,
Kvasir/U-Net MC improves from 0.580 to 0.868 error capture at a 20\% budget.

These results should still be interpreted cautiously. On AURC, plain uncertainty
is ahead in 3/18 combinations, so a learned policy is not uniformly better than
the raw score. Validation size also matters: DRIVE has only four validation
images for fitting the deployable learned head. Although this provides many
pixels, it is image-starved and cannot show that the learned policy has seen
enough anatomical variation. Learned deferral should therefore be reported with
its fitting data and evaluated on the target distribution.

\subsection{Review assumptions can change the winner}
\label{sec:results_review_assumptions}

\begin{figure}[t]
\centering
\includegraphics[width=\linewidth]{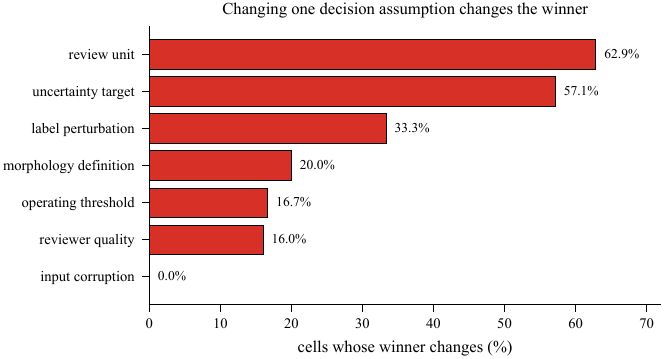}
\caption{\textbf{Decision-layer sensitivity on the DRIVE anchor.}
Fraction of tested settings in which changing one review assumption changes the
preferred estimator. Predictions are held fixed, so reversals isolate the
decision layer. These are finite-grid sensitivity summaries, not population
estimates.}
\label{fig:decision-reversal}
\end{figure}

Fixed-budget error capture assumes that deferred units can be corrected. This is
an analytical review proxy, not a reader-study measurement, so we audit whether
conclusions are stable under alternative review assumptions. They are not always
stable. Changing the review unit flips the estimator ranking in 62.9\% of tested
settings, changing the uncertainty target flips it in 57.1\%, and changing the
cost/utility assumption changes the preferred method across much of the cost
grid. Threshold choice, label perturbation, morphology definition, and imperfect
reviewer models also produce ranking reversals, whereas the tested corruption
grid gives no MC--TTA reversal.

This makes the review unit part of the evaluation rather than an implementation
detail. Pixel-level deferral is analytically convenient, but review is usually
organised over patches, components, or images. In the region-unit analysis,
DRIVE connected-component deferral captures about 0.88 of errors at a 20\%
budget with accepted Dice near 0.94, while fixed blocks capture much less. On
Kvasir/U-Net, changing only the review unit moves MC from 0.462 error capture
with $16\times16$ blocks to 0.804 with top-$k$ 16-pixel patches. \Cref{fig:decision-reversal} summarizes these reversals.

\subsection{Boundary failures are partly annotation ambiguity}
\label{sec:results_stratified}

Pooled AUROC hides spatial failure modes. In the nine-cell matrix, boundary-band
AUROC is consistently below global AUROC, and thin-structure detection is often
weaker still. The DRIVE second-observer analysis shows why this must be stated
carefully. Only 5.3\% of field-of-view pixels are disputed by the two observers,
but 91.7\% of that disagreement lies inside a two-pixel boundary band. On
observer-disagreement pixels, every estimator is near chance AUROC. Uncertainty
can mark unstable regions, but it cannot adjudicate between two plausible
annotations.

Annotation ambiguity does not fully explain the boundary and thin-structure
weakness. Restricting to observer-agreement pixels removes only a minority of
the boundary and thin penalties. Under the observer-2 analysis, MC and TTA are
statistically indistinguishable globally, while TTA's advantage is large and
significant on thin structures. Boundary conclusions are sensitive to the chosen
morphological radius, whereas the thin-structure advantage is stable across
radii. We therefore state the structural result narrowly: uncertainty is weakest
near ambiguous boundaries and thin structures, and this weakness is partly, but
not entirely, explained by annotation disagreement.

\begin{figure*}[t]
  \centering
  \includegraphics[width=\textwidth]{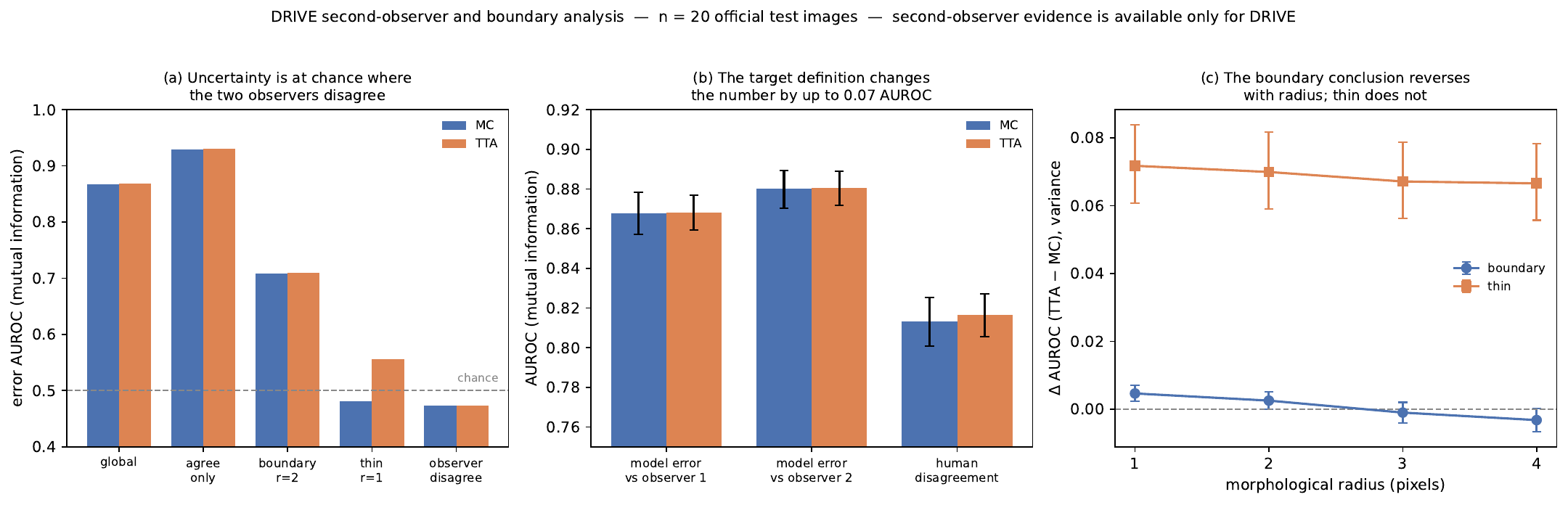}
  \caption{DRIVE second-observer analysis. Uncertainty is near chance on pixels
  where the two observers disagree; changing the target definition changes the
  AUROC; and boundary conclusions depend on the morphological radius, while
  thin-structure effects are more stable.}
  \label{fig:drive_observer}
\end{figure*}

\subsection{Policy transfer remains partial}
\label{sec:results_transfer}

\begin{table}[t]
\centering
\small
\setlength{\tabcolsep}{4pt}
\caption{\textbf{Policy, transfer, and annotation audits.} Compact summary of
decision-layer results. Learned-deferral counts use the deployable
validation-fit protocol; transfer uses source-validation fitting and held-out
target tests.}
\label{tab:policy_annotation_audit}
\resizebox{\columnwidth}{!}{%
\begin{tabular}{lp{0.62\columnwidth}}
\toprule
Audit & Main result \\
\midrule
Learned deferral &
Capt.@20\% improves in 16/18 cell--estimator settings; DRIVE uses only 4
validation images for fitting. \\
Transfer vs raw target $u$ &
8/18 significantly better, 7/18 significantly worse, 3/18 n.s. \\
Transfer vs in-domain refit &
4/18 significantly better, 9/18 significantly worse, 5/18 n.s.; refitting is
usually safer. \\
DRIVE second observer &
Global MC--TTA differences are n.s.; TTA gains concentrate on thin structures
($+0.072$ variance, $+0.075$ MI). \\
Human-disagreement target &
MI AUROC is 0.813 for MC and 0.817 for TTA, but all scores are near chance on
disputed pixels. \\
\bottomrule
\end{tabular}%
}
\end{table}

For policy transfer, source policies are fit only on source validation data and
evaluated on held-out target test images, with zero source-validation/target-test
overlap. As summarized in \cref{tab:policy_annotation_audit}, transfer can
improve over raw target uncertainty in some pairs, but is usually worse than an
in-domain refit. Review policies should therefore be evaluated, and where
possible fit or calibrated, on the target distribution.

\section{Discussion}
\label{sec:discussion}

The results support a simple conclusion: segmentation uncertainty should be
evaluated by the review decisions it enables, not by the score map alone. Under
matched budgets and canonical multi-layer MC Dropout, TTA gives a stronger
error-ranking signal across our MC--TTA comparisons. However, the broader
benchmark does not support a universal estimator ranking: deep ensembles can
outperform TTA on some datasets, probability-derived scores remain competitive,
learned deferral is not uniformly better than raw uncertainty, and review
assumptions can change the preferred method. These patterns would be obscured by
reporting only calibration error or a single pooled Unc-AUROC.

Simple baselines are also stronger than they may appear. In binary segmentation,
entropy, margin, inverse confidence, distance from $1/2$, and maximum-softmax
confidence induce the same pixel ranking and therefore the same deferred set at
a fixed budget. More generally, any strictly monotone transformation of a review
score preserves coverage-based deferral, leaving the deferred set,
risk--coverage curve, and AURC unchanged. Thus, better calibrated scores or
alternative softmax-derived scores do not necessarily imply better review
decisions.

The review unit and review assumptions are part of the evaluation. Pixel-level
uncertainty is useful for controlled comparison, but practical review is usually
organised over regions, patches, connected components, or whole images. Our
region-unit and sensitivity analyses show that changing the review unit,
uncertainty target, correction assumption, or cost model can substantially
change the ranking. 
\Cref{tab:reporting_checklist} summarizes the corresponding minimum reporting
fields. Learned policies should be compared against raw target uncertainty and,
where possible, fit or calibrated on target-domain validation data.

\textbf{Recommended use.}
For a new segmentation setting, we recommend reporting: (i) a probability-derived
baseline, (ii) ERR@$\beta$ and risk--coverage under the intended review budget,
(iii) structural stratification for boundary and thin-structure errors, and
(iv) at least one region-unit proxy matched to the expected review workflow.
Learned policies should be compared against raw target uncertainty and, where
possible, fit or calibrated on target-domain validation data.

\begin{table}[t]
\centering
\small
\setlength{\tabcolsep}{4pt}
\caption{\textbf{Minimum reporting checklist for decision-aware uncertainty
evaluation.} The estimator alone is insufficient; the reported result should
identify the full configuration that turns uncertainty into a review decision.}
\label{tab:reporting_checklist}
\resizebox{\columnwidth}{!}{%
\begin{tabular}{lp{0.62\columnwidth}}
\toprule
Coordinate & What should be reported \\
\midrule
Estimator & MC/TTA/ensemble/probability score and uncertainty functional. \\
Compute & Number of passes or ensemble members; matched or deliberately swept. \\
Backbone & Segmentation checkpoint and whether the comparator changes it. \\
Policy & Raw ranking, adaptive threshold, learned head, or conformal rule. \\
Fitting data & Validation images used for policy fitting; no test-label fitting. \\
Review unit & Pixel, block, patch, connected component, region, or image. \\
Target & All errors, false positives, false negatives, boundary/thin errors, or disagreement. \\
Review model & Perfect/imperfect correction assumption and cost or utility model. \\
Statistics & Paired held-out-image inference and multiplicity family. \\
\bottomrule
\end{tabular}%
}
\end{table}

\section{Limitations}
\label{sec:limitations}

\noindent\textbf{Scope and generality.}
The benchmark spans four public datasets, selected backbones, and a controlled
set of uncertainty estimators and policies. Matched-budget controls,
complementary-architecture checks, and dropout-placement experiments strengthen
the MC--TTA comparison, but they do not constitute a population-level sample of
all segmentation tasks. We therefore treat cross-dataset counts as
benchmark evidence rather than universal claims.

\noindent\textbf{Compute and estimator families.}
MC Dropout, TTA, and deep ensembles all add inference or training cost. Our main
MC--TTA comparison uses matched inference budgets, but unequal-budget stress
tests show that giving MC many more stochastic samples can produce ties or
reversals. We therefore do not claim a compute-independent ranking or optimal
pass count; wall-clock cost depends on hardware, batching, augmentation
implementation, and ensemble training.

\noindent\textbf{Policy and validation limits.}
Learned deferral heads are fit only on validation data and are never trained on
test labels, but their reliability depends on validation-set size and diversity.
DRIVE is especially limited, with only four validation images for the deployable
learned head. Learned policies transfer only partially, so they should be
evaluated, and where possible refit, on the target distribution.

\noindent\textbf{Annotation and clinical generalisation.}
Second-observer analysis is available only for DRIVE, so annotation-disagreement
claims should not be generalized to the other datasets without additional
labels. Fixed-budget deferral is also an analytical proxy for expert review
under the assumption that deferred units can be corrected. We do not conduct a
reader study, measure clinical time, or perform prospective deployment. The paper
should therefore be read as an evaluation protocol and empirical audit, not as a
claim of clinical readiness.

\section{Conclusion}
\label{sec:conclusion}

We presented a decision-aware evaluation protocol for medical segmentation
uncertainty. The central claim is that uncertainty should be judged by the review
decisions it enables under explicit budgets, not by calibration or
error-detection scores alone. By separating uncertainty estimation from deferral
policy, the protocol compares MC Dropout, TTA, deep ensembles,
probability-derived scores, and learned review policies as
estimator--policy--review configurations. Within our benchmark, matched-budget
TTA gives a stronger error-ranking signal than canonical multi-layer MC Dropout,
but the broader evaluation does not support a universal estimator ordering:
deep ensembles can outperform TTA on some datasets, probability-derived scores
remain competitive, learned deferral is not uniformly better than raw
uncertainty, and review assumptions can change the preferred method. Structural
and second-observer analyses further show that boundary and thin-structure
failures are partly tied to annotation ambiguity. These findings argue against
choosing an uncertainty estimator from a single score-level metric. Instead,
uncertainty should be audited on the target distribution, at the intended review
budget, and with review units and assumptions that reflect the workflow. We will
release the protocol, code, artifacts, and summaries to support reproducible
decision-aware evaluation on new datasets, backbones, estimators, and review
policies.

% Flush the pending full-width qualitative figure before the bibliography so it
% is placed within the body rather than drifting into the references.

{
    \small
    \bibliographystyle{ieeenat_fullname}
    \bibliography{main}

@String(ICCV= {Int. Conf. Comput. Vis.})

@String(ECCV= {Eur. Conf. Comput. Vis.})

@String(ICLR = {Int. Conf. Learn. Represent.})

@String(CVPRW= {IEEE Conf. Comput. Vis. Pattern Recog. Worksh.})

@inproceedings{gal2016dropout,
  title={Dropout as a {Bayesian} approximation: Representing model uncertainty in deep learning},
  author={Gal, Yarin and Ghahramani, Zoubin},
  booktitle={International Conference on Machine Learning},
  pages={1050--1059},
  year={2016}
}

@inproceedings{medklip,
  title={Medklip: Medical knowledge enhanced language-image pre-training for x-ray diagnosis},
  author={Wu, Chaoyi and Zhang, Xiaoman and Zhang, Ya and Wang, Yanfeng and Xie, Weidi},
  booktitle={2023 IEEE/CVF international conference on computer vision (ICCV)},
  pages={21315--21326},
  year={2023},
  organization={IEEE}
}

@article{liu2024g2d,
  title={G2d: From global to dense radiography representation learning via vision-language pre-training},
  author={Liu, Che and Ouyang, Cheng and Cheng, Sibo and Shah, Anand and Bai, Wenjia and Arcucci, Rossella},
  journal={Advances in Neural Information Processing Systems},
  volume={37},
  pages={14751--14773},
  year={2024}
}

@inproceedings{lakshminarayanan2017simple,
  title={Simple and scalable predictive uncertainty estimation using deep ensembles},
  author={Lakshminarayanan, Balaji and Pritzel, Alexander and Blundell, Charles},
  booktitle={Advances in Neural Information Processing Systems},
  pages={6402--6413},
  year={2017}
}

@inproceedings{gowda2024masks,
  title={Masks and manuscripts: Advancing medical pre-training with end-to-end masking and narrative structuring},
  author={Gowda, Shreyank N and Clifton, David A},
  booktitle={International Conference on Medical Image Computing and Computer-Assisted Intervention},
  pages={426--436},
  year={2024},
  organization={Springer}
}

@inproceedings{cc-sam,
  title={Cc-sam: Sam with cross-feature attention and context for ultrasound image segmentation},
  author={Gowda, Shreyank N and Clifton, David A},
  booktitle={European Conference on Computer Vision},
  pages={108--124},
  year={2024},
  organization={Springer}
}

@article{mehrtash2020confidence,
  title={Confidence calibration and predictive uncertainty estimation for deep medical image segmentation},
  author={Mehrtash, Alireza and Wells, William M and Tempany, Clare M and Abolmaesumi, Purang and Kapur, Tina},
  journal={IEEE Transactions on Medical Imaging},
  volume={39},
  number={12},
  pages={3868--3878},
  year={2020},
  doi={10.1109/TMI.2020.3006437}
}

@inproceedings{ronneberger2015unet,
  title={{U-Net}: Convolutional networks for biomedical image segmentation},
  author={Ronneberger, Olaf and Fischer, Philipp and Brox, Thomas},
  booktitle={International Conference on Medical Image Computing and Computer-Assisted Intervention},
  pages={234--241},
  year={2015}
}

@inproceedings{zhou2018unetplusplus,
  title={{UNet++}: A nested {U-Net} architecture for medical image segmentation},
  author={Zhou, Zongwei and Siddiquee, Md Mahfuzur Rahman and Tajbakhsh, Nima and Liang, Jianming},
  booktitle={Deep Learning in Medical Image Analysis},
  pages={3--11},
  year={2018}
}

@article{isensee2021nnunet,
  title={{nnU-Net}: a self-configuring method for deep learning-based biomedical image segmentation},
  author={Isensee, Fabian and Jaeger, Paul F. and Kohl, Simon A. A. and Petersen, Jens and Maier-Hein, Klaus H.},
  journal={Nature Methods},
  volume={18},
  number={2},
  pages={203--211},
  year={2021},
  doi={10.1038/s41592-020-01008-z}
}

@article{chen2021transunet,
  title={{TransUNet}: Transformers make strong encoders for medical image segmentation},
  author={Chen, Jieneng and Lu, Yongyi and Yu, Qihang and Luo, Xiangde and Adeli, Ehsan and Wang, Yan and Lu, Le and Yuille, Alan L and Zhou, Yuyin},
  journal={arXiv preprint arXiv:2102.04306},
  year={2021}
}

@inproceedings{kirillov2023segment,
  title={Segment Anything},
  author={Kirillov, Alexander and Mintun, Eric and Ravi, Nikhila and Mao, Hanzi and Rolland, Chloe and Gustafson, Laura and Xiao, Tete and Whitehead, Spencer and Berg, Alexander C. and Lo, Wan-Yen and Doll{\'a}r, Piotr and Girshick, Ross},
  booktitle={IEEE/CVF International Conference on Computer Vision},
  pages={3992--4003},
  year={2023},
  doi={10.1109/ICCV51070.2023.00371}
}

@article{ma2024medsam,
  title={Segment anything in medical images},
  author={Ma, Jun and He, Yuting and Li, Feifei and Han, Lin and You, Chenyu and Wang, Bo},
  journal={Nature Communications},
  volume={15},
  pages={654},
  year={2024},
  doi={10.1038/s41467-024-44824-z}
}

@article{ulmer2023prior,
  title={Prior and posterior networks: A survey on evidential deep learning methods for uncertainty estimation},
  author={Ulmer, Dennis and Hardmeier, Christian and Frellsen, Jes},
  journal={Transactions on Machine Learning Research (TMLR)},
  year={2023}
}

@inproceedings{ayhan2018test,
  title={Test-time data augmentation for estimation of heteroscedastic aleatoric uncertainty in deep neural networks},
  author={Ayhan, Murat Seckin and Berens, Philipp},
  booktitle={International Conference on Medical Imaging with Deep Learning},
  year={2018}
}

@article{wang2019aleatoric,
  title={Aleatoric uncertainty estimation with test-time augmentation for medical image segmentation with convolutional neural networks},
  author={Wang, Guotai and Li, Wenqi and Aertsen, Michael and Deprest, Jan and Ourselin, Sebastien and Vercauteren, Tom},
  journal={Neurocomputing},
  volume={338},
  pages={34--45},
  year={2019},
  doi={10.1016/j.neucom.2019.01.103}
}

@article{moshkov2020tta,
  title={Test-time augmentation for deep learning-based cell segmentation on microscopy images},
  author={Moshkov, Nikita and Mathe, Botond and Kertesz-Farkas, Attila and Hollandi, Reka and Horvath, Peter},
  journal={Scientific Reports},
  volume={10},
  pages={5068},
  year={2020},
  doi={10.1038/s41598-020-61808-3}
}

@inproceedings{czolbe2021segmentation,
  title={Is segmentation uncertainty useful?},
  author={Czolbe, Steffen and Arnavaz, Kasra and Krause, Oswin and Feragen, Aasa},
  booktitle={Information Processing in Medical Imaging (IPMI)},
  pages={715--726},
  year={2021},
  doi={10.1007/978-3-030-78191-0_55}
}

@inproceedings{band2021benchmarking,
  title={Benchmarking {Bayesian} deep learning on diabetic retinopathy detection tasks},
  author={Band, Neil and Rudner, Tim G. J. and Feng, Qixuan and Filos, Angelos and Nado, Zachary and Dusenberry, Michael W. and Jerfel, Ghassen and Tran, Dustin and Gal, Yarin},
  booktitle={Advances in Neural Information Processing Systems (NeurIPS), Datasets and Benchmarks Track},
  year={2021}
}

@inproceedings{guo2017calibration,
  title={On calibration of modern neural networks},
  author={Guo, Chuan and Pleiss, Geoff and Sun, Yu and Weinberger, Kilian Q},
  booktitle={International Conference on Machine Learning},
  pages={1321--1330},
  year={2017}
}

@inproceedings{geifman2017selective,
  title={Selective classification for deep neural networks},
  author={Geifman, Yonatan and El-Yaniv, Ran},
  booktitle={Advances in Neural Information Processing Systems},
  pages={4878--4887},
  year={2017}
}

@article{vickers2006decision,
  title={Decision curve analysis: a novel method for evaluating prediction models},
  author={Vickers, Andrew J. and Elkin, Elena B.},
  journal={Medical Decision Making},
  volume={26},
  number={6},
  pages={565--574},
  year={2006},
  publisher={SAGE Publications},
  doi={10.1177/0272989X06295361}
}

@article{vickers2019dca,
  title={A simple, step-by-step guide to interpreting decision curve analysis},
  author={Vickers, Andrew J. and van Calster, Ben and Steyerberg, Ewout W.},
  journal={Diagnostic and Prognostic Research},
  volume={3},
  number={1},
  pages={18},
  year={2019},
  publisher={Springer},
  doi={10.1186/s41512-019-0064-7}
}

@article{chow1970optimum,
  title={On optimum recognition error and reject tradeoff},
  author={Chow, Chi-Keung},
  journal={IEEE Transactions on Information Theory},
  volume={16},
  number={1},
  pages={41--46},
  year={1970}
}

@inproceedings{verma2022calibrated,
  title={Calibrated learning to defer with one-vs-all classifiers},
  author={Verma, Rajeev and Nalisnick, Eric},
  booktitle={International Conference on Machine Learning (ICML)},
  series={Proceedings of Machine Learning Research},
  volume={162},
  pages={22184--22202},
  year={2022}
}

@inproceedings{mozannar2023who,
  title={Who should predict? {Exact} algorithms for learning to defer to humans},
  author={Mozannar, Hussein and Lang, Hunter and Wei, Dennis and Sattigeri, Prasanna and Das, Subhro and Sontag, David},
  booktitle={International Conference on Artificial Intelligence and Statistics (AISTATS)},
  series={Proceedings of Machine Learning Research},
  volume={206},
  pages={10520--10545},
  year={2023}
}

@inproceedings{angelopoulos2024conformal,
  title={Conformal risk control},
  author={Angelopoulos, Anastasios N. and Bates, Stephen and Fisch, Adam and Lei, Lihua and Schuster, Tal},
  booktitle={International Conference on Learning Representations (ICLR)},
  year={2024}
}

@article{staal2004drive,
  title={Ridge-based vessel segmentation in color images of the retina},
  author={Staal, Joes and Abramoff, Michael D and Niemeijer, Meindert and Viergever, Max A and Van Ginneken, Bram},
  journal={IEEE Transactions on Medical Imaging},
  volume={23},
  number={4},
  pages={501--509},
  year={2004},
  doi={10.1109/TMI.2004.825627}
}

@inproceedings{ravi2025sam2,
  title={{SAM} 2: Segment Anything in Images and Videos},
  author={Ravi, Nikhila and Gabeur, Valentin and Hu, Yuan-Ting and Hu, Ronghang and Ryali, Chaitanya and Ma, Tengyu and Khedr, Haitham and R{\"a}dle, Roman and Rolland, Chloe and Gustafson, Laura and Mintun, Eric and Pan, Junting and Alwala, Kalyan Vasudev and Carion, Nicolas and Wu, Chao-Yuan and Girshick, Ross and Doll{\'a}r, Piotr and Feichtenhofer, Christoph},
  booktitle={International Conference on Learning Representations (ICLR)},
  year={2025}
}

@inproceedings{mossina2024conformal,
  title={Conformal Semantic Image Segmentation: Post-hoc Quantification of Predictive Uncertainty},
  author={Mossina, Luca and Dalmau, Joseba and And{\'e}ol, L{\'e}o},
  booktitle={IEEE/CVF Conference on Computer Vision and Pattern Recognition Workshops (CVPRW)},
  year={2024}
}

@article{jin2022fives,
  title={{FIVES}: A Fundus Image Dataset for Artificial Intelligence based Vessel Segmentation},
  author={Jin, Kai and Huang, Xingru and Zhou, Jingxing and Li, Yunxiang and Yan, Yan and Sun, Yibao and Zhang, Qianni and Wang, Yaqi and Ye, Juan},
  journal={Scientific Data},
  volume={9},
  number={1},
  pages={475},
  year={2022},
  publisher={Nature Publishing Group},
  doi={10.1038/s41597-022-01564-3}
}

@article{codella2019isic,
  title={Skin Lesion Analysis Toward Melanoma Detection 2018: A Challenge Hosted by the International Skin Imaging Collaboration ({ISIC})},
  author={Codella, Noel and Rotemberg, Veronica and Tschandl, Philipp and Celebi, M Emre and Dusza, Stephen and Gutman, David and Helba, Brian and Kalloo, Aadi and Liopyris, Konstantinos and Marchetti, Michael and Kittler, Harald and Halpern, Allan},
  journal={arXiv preprint arXiv:1902.03368},
  year={2019}
}

@inproceedings{jha2020kvasir,
  title={{Kvasir-SEG}: A Segmented Polyp Dataset},
  author={Jha, Debesh and Smedsrud, Pia H and Riegler, Michael A and Halvorsen, P{\aa}l and de Lange, Thomas and Johansen, Dag and Johansen, H{\aa}vard D},
  booktitle={International Conference on Multimedia Modeling (MMM)},
  pages={451--462},
  year={2020},
  organization={Springer},
  doi={10.1007/978-3-030-37734-2_37}
}

@inproceedings{chen2018deeplabv3plus,
  title={Encoder-Decoder with Atrous Separable Convolution for Semantic Image Segmentation},
  author={Chen, Liang-Chieh and Zhu, Yukun and Papandreou, George and Schroff, Florian and Adam, Hartwig},
  booktitle={European Conference on Computer Vision (ECCV)},
  pages={833--851},
  year={2018},
  doi={10.1007/978-3-030-01234-2_49}
}

@inproceedings{ovadia2019trust,
  title={Can you trust your model's uncertainty? {Evaluating} predictive uncertainty under dataset shift},
  author={Ovadia, Yaniv and Fertig, Emily and Ren, Jie and Nado, Zachary and Sculley, David and Nowozin, Sebastian and Dillon, Joshua V. and Lakshminarayanan, Balaji and Snoek, Jasper},
  booktitle={Advances in Neural Information Processing Systems},
  year={2019}
}

@article{bates2021rcps,
  title={Distribution-free, risk-controlling prediction sets},
  author={Bates, Stephen and Angelopoulos, Anastasios and Lei, Lihua and Malik, Jitendra and Jordan, Michael I.},
  journal={Journal of the ACM},
  volume={68},
  number={6},
  pages={1--34},
  year={2021}
}

@inproceedings{corbiere2019addressing,
  title={Addressing failure prediction by learning model confidence},
  author={Corbi{\`e}re, Charles and Thome, Nicolas and Bar-Hen, Avner and Cord, Matthieu and P{\'e}rez, Patrick},
  booktitle={Advances in Neural Information Processing Systems},
  year={2019}
}

@inproceedings{hendrycks2017baseline,
  title={A baseline for detecting misclassified and out-of-distribution examples in neural networks},
  author={Hendrycks, Dan and Gimpel, Kevin},
  booktitle={International Conference on Learning Representations (ICLR)},
  year={2017}
}

@inproceedings{jungo2019assessing,
  title={Assessing reliability and challenges of uncertainty estimations for medical image segmentation},
  author={Jungo, Alain and Reyes, Mauricio},
  booktitle={International Conference on Medical Image Computing and Computer-Assisted Intervention (MICCAI)},
  pages={48--56},
  year={2019}
}

@inproceedings{sensoy2018evidential,
  title={Evidential deep learning to quantify classification uncertainty},
  author={Sensoy, Murat and Kaplan, Lance and Kandemir, Melih},
  booktitle={Advances in Neural Information Processing Systems},
  year={2018}
}

@article{mazurowski2023segment,
  title={Segment anything model for medical image analysis: An experimental study},
  author={Mazurowski, Maciej A. and Dong, Haoyu and Gu, Hanxue and Yang, Jichen and Konz, Nicholas and Zhang, Yixin},
  journal={Medical Image Analysis},
  volume={89},
  pages={102918},
  year={2023}
}

@inproceedings{shanmugam2021better,
  title={Better aggregation in test-time augmentation},
  author={Shanmugam, Divya and Blalock, Davis and Balakrishnan, Guha and Guttag, John},
  booktitle={Proceedings of the IEEE/CVF International Conference on Computer Vision (ICCV)},
  pages={1214--1223},
  year={2021}
}

@article{nair2020exploring,
  title={Exploring uncertainty measures in deep networks for multiple sclerosis lesion detection and segmentation},
  author={Nair, Tanya and Precup, Doina and Arnold, Douglas L. and Arbel, Tal},
  journal={Medical Image Analysis},
  volume={59},
  pages={101557},
  year={2020}
}

@article{angelopoulos2021gentle,
  title={A gentle introduction to conformal prediction and distribution-free uncertainty quantification},
  author={Angelopoulos, Anastasios N. and Bates, Stephen},
  journal={arXiv preprint arXiv:2107.07511},
  year={2021}
}

@inproceedings{mozannar2020consistent,
  title={Consistent estimators for learning to defer to an expert},
  author={Mozannar, Hussein and Sontag, David},
  booktitle={International Conference on Machine Learning (ICML)},
  pages={7076--7087},
  year={2020}
}

@inproceedings{geifman2019selectivenet,
  title={{SelectiveNet}: A deep neural network with an integrated reject option},
  author={Geifman, Yonatan and El-Yaniv, Ran},
  booktitle={International Conference on Machine Learning (ICML)},
  pages={2151--2159},
  year={2019}
}
}
\clearpage

This supplement provides the audit trail for the main paper. It reports the
statistical procedure, dataset splits, matched-budget MC Dropout--TTA
comparisons, robustness checks, deep-ensemble comparisons, learned-deferral and
transfer analyses, review-assumption sensitivity, annotation-disagreement
analysis, and reproducibility details. Unless stated otherwise, comparisons are
paired at the held-out-image level. Null, reversed, blocked, and
dataset-limited findings are retained and labelled as such.

\subsection{Statistical procedure and multiplicity}
\label{supp:stats}

For paired image-level scores $a_i,b_i$, the estimand is

\begin{equation}
  \Delta = n^{-1}\sum_i(a_i-b_i).  
\end{equation}

We report percentile bootstrap confidence intervals over images and two-sided
paired sign-flip randomisation tests, followed by Benjamini--Hochberg (BH)
correction within each declared comparison family. Policy-transfer analyses also
report exact sign tests. We do not form confidence intervals by treating
dataset--backbone cells as IID replicates.

The main comparison families are matched-budget MC Dropout versus TTA,
stochastic replication, complementary architecture, dropout placement,
dropout-rate and training-seed sensitivity, unequal-budget comparisons,
deep-ensemble comparisons, learned-deferral fixed-budget tests, policy transfer,
review-assumption sensitivity, and second-observer analyses. Cross-cell counts
are used descriptively; image-level paired data are used whenever available.

\subsection{Dataset splits and analysis coverage}
\label{supp:splits}

\begin{table*}[t]
\centering
\scriptsize
\setlength{\tabcolsep}{2.2pt}
\caption{\textbf{Exact split and analysis counts.} DRIVE uses the official 20-image test half but a self-defined 16/4 train/validation subdivision. FIVES, ISIC, and Kvasir use self-defined 65/15/20 splits (seed 42). Zero denotes an analysis not run for that architecture arm.}
\label{tab:supp-splits-full}
\resizebox{\textwidth}{!}{%
\begin{tabular}{lllrrrlrrrrr}
\toprule
Dataset & Architecture & Wave & Train & Val & Test & Split status & MC & TTA & Ens. & Head fit & Head eval \\
\midrule
DRIVE & U-Net & canonical + Waves A/C/D/E/F & 16 & 4 & 20 & official test, self-defined train/val & 20 & 20 & 20 & 4 & 20 \\
DRIVE & MiT-B2 TransUNet & Wave B (complementary architecture) & 16 & 4 & 20 & official test, self-defined train/val & 20 & 20 & 0 & 0 & 0 \\
FIVES & MiT-B2 encoder + U-Net decoder & canonical + Waves A/C/D/E/F & 520 & 120 & 160 & self-defined & 160 & 160 & 160 & 120 & 160 \\
FIVES & U-Net & Wave B (complementary architecture) & 520 & 120 & 160 & self-defined & 160 & 160 & 0 & 0 & 0 \\
ISIC & U-Net & canonical + Waves A/C/D/E/F & 1686 & 389 & 519 & self-defined & 519 & 519 & 519 & 389 & 519 \\
ISIC & MiT-B2 TransUNet & Wave B (complementary architecture) & 1686 & 389 & 519 & self-defined & 519 & 519 & 0 & 0 & 0 \\
Kvasir & U-Net & canonical + Waves A/C/D/E/F & 650 & 150 & 200 & self-defined & 200 & 200 & 200 & 150 & 200 \\
Kvasir & MiT-B2 TransUNet & Wave B (complementary architecture) & 650 & 150 & 200 & self-defined & 200 & 200 & 0 & 0 & 0 \\
\bottomrule
\end{tabular}}
\end{table*}

DRIVE uses the official 20-image test set with first-observer annotations and a
self-defined train/validation split from the official training half. FIVES,
ISIC, and Kvasir use self-defined 65/15/20 splits with a fixed seed. These
splits are suitable for paired uncertainty comparisons within this protocol, but
should not be interpreted as canonical benchmark partitions for absolute
segmentation-performance comparison. The split hash, test-ID hash, segmentation
threshold, and field-of-view rule are pinned in the released manifests.

\subsection{Matched-budget MC Dropout versus TTA}
\label{supp:matched}

The primary MC--TTA comparison uses canonical multi-layer MC Dropout with
normalisation layers frozen at test time. MC Dropout and TTA are evaluated under
matched inference budgets $T\in{2,4,6}$ and on the same held-out images within
each dataset. We report both predictive variance and mutual information as
uncertainty functionals.

% Generated by wacv_r2_final/scripts/run_05_build_tables.py. Do not edit by hand.
\begin{table}[t]
\centering
\small
\setlength{\tabcolsep}{2pt}
\caption{\textbf{Matched inference budgets, canonical multi-layer MC Dropout.}
Mean image-level error-detection AUROC; MC and TTA draw the same $T$ passes from the
same checkpoint on the same images. $\Delta$ is the paired per-image mean TTA${-}$MC;
\textsuperscript{*} marks $q<0.05$ after BH correction across this family of 24.
TTA leads in all 24 and every bootstrap interval excludes zero.}
\label{tab:matched-budget}
\begin{tabular}{llcccccc}
\toprule
 & & \multicolumn{2}{c}{$T{=}2$} & \multicolumn{2}{c}{$T{=}4$} & \multicolumn{2}{c}{$T{=}6$} \\
\cmidrule(lr){3-4}\cmidrule(lr){5-6}\cmidrule(lr){7-8}
Dataset & Func. & MC & $\Delta$ & MC & $\Delta$ & MC & $\Delta$ \\
\midrule
DRIVE & Var & 0.805 & +0.017\textsuperscript{*} & 0.831 & +0.019\textsuperscript{*} & 0.839 & +0.019\textsuperscript{*} \\
DRIVE & MI & 0.807 & +0.016\textsuperscript{*} & 0.843 & +0.017\textsuperscript{*} & 0.852 & +0.016\textsuperscript{*} \\
FIVES & Var & 0.829 & +0.004\textsuperscript{*} & 0.844 & +0.008\textsuperscript{*} & 0.852 & +0.007\textsuperscript{*} \\
FIVES & MI & 0.837 & +0.010\textsuperscript{*} & 0.853 & +0.014\textsuperscript{*} & 0.859 & +0.014\textsuperscript{*} \\
ISIC & Var & 0.761 & +0.056\textsuperscript{*} & 0.793 & +0.060\textsuperscript{*} & 0.803 & +0.064\textsuperscript{*} \\
ISIC & MI & 0.776 & +0.058\textsuperscript{*} & 0.812 & +0.052\textsuperscript{*} & 0.821 & +0.053\textsuperscript{*} \\
Kvasir & Var & 0.783 & +0.057\textsuperscript{*} & 0.817 & +0.064\textsuperscript{*} & 0.832 & +0.065\textsuperscript{*} \\
Kvasir & MI & 0.802 & +0.059\textsuperscript{*} & 0.847 & +0.052\textsuperscript{*} & 0.864 & +0.047\textsuperscript{*} \\
\bottomrule
\end{tabular}
\end{table}

Across four datasets, three budgets, and two uncertainty functionals, TTA has
higher image-level error-detection AUROC than MC Dropout in all 24 matched
comparisons, and every bootstrap interval excludes zero after BH correction.
This establishes the main MC--TTA result under matched compute and a stronger
canonical MC Dropout configuration.

\subsection{Robustness checks}
\label{supp:robustness}

The matched-budget result is stress-tested along several axes: independent MC
stochastic realisations, all TTA transform subsets, complementary architecture,
dropout placement, dropout rate, training seed, and unequal compute budgets.

% Generated by wacv_r2_final/scripts/run_05_build_tables.py. Do not edit by hand.
\begin{table}[t]
\centering
\footnotesize
\setlength{\tabcolsep}{3pt}
\caption{\textbf{Robustness across every configuration axis raised in review.}
Each row is an independent family of paired image-level comparisons with its own BH
correction; ``TTA'' counts positive point estimates and ``BH'' counts $q<0.05$. The matched-budget, replication, architecture, and placement families are uniformly positive;
the two DRIVE-only sweeps show smaller effects, non-significant cells, and three tiny seed reversals.}
\label{tab:robustness}
\resizebox{\linewidth}{!}{%
\begin{tabular}{lrrrr}
\toprule
Axis & TTA & BH & mean $\Delta$ & range $\Delta$ \\
\midrule
Matched budget, anchor (4 ds) & 24/24 & 24/24 & +0.035 & [+0.004, +0.065] \\
10 MC realizations (4 ds) & 24/24 & 24/24 & +0.033 & [+0.003, +0.063] \\
Complementary backbone (4 ds) & 24/24 & 24/24 & +0.063 & [+0.015, +0.133] \\
Dropout placement (4 ds) & 72/72 & 72/72 & +0.075 & [+0.004, +0.250] \\
Dropout rate $p$ (DRIVE) & 24/24 & 18/24 & +0.015 & [+0.003, +0.038] \\
Training seed (DRIVE) & 15/18 & 10/18 & +0.009 & [-0.004, +0.020] \\
\bottomrule
\end{tabular}}
\end{table}

With ten independent complete MC sequences, canonical TTA exceeds every MC
realisation in all matched dataset--budget--functional comparisons. Enumerating
TTA subsets gives the same overall direction: 2,342/2,480 subset--realisation
pairings favour TTA, and even the weakest TTA subset exceeds the strongest MC
realisation in 21/24 cells. The three exceptions are low-budget FIVES settings.

\begin{figure*}[t]
\centering
\includegraphics[width=0.88\textwidth]{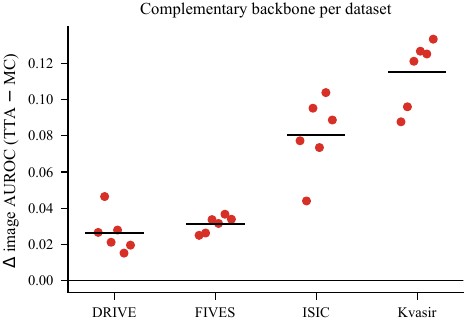}
\caption{\textbf{Complementary architecture audit.} DRIVE, ISIC, and Kvasir
are evaluated with a MiT-B2 TransUNet variant, while FIVES is evaluated with a
U-Net variant. The paired TTA--MC effect remains positive in all
budget--functional comparisons after the architecture swap.}
\label{fig:supp-architecture}
\end{figure*}

The complementary-architecture experiment tests whether the MC--TTA result is
tied to one backbone family. Within each arm, MC and TTA share the same trained
checkpoint, split, threshold, field-of-view rule, and pass budget. TTA remains
above MC in all matched complementary-architecture comparisons, but we restrict
the conclusion to the architecture families tested.

Dropout placement also matters. Head-only MC is weaker than canonical
multi-layer MC in all matched comparisons, while deep/bottleneck-only MC is often
stronger than canonical MC. Despite this movement in the comparator, TTA remains
above MC within each matched placement arm. Dropout-rate and training-seed
sweeps are DRIVE-only sensitivity analyses: they preserve the overall direction
but show smaller effects, non-significant cells, and small low-budget reversals.

\begin{figure*}[t]
\centering
\includegraphics[width=0.88\textwidth]{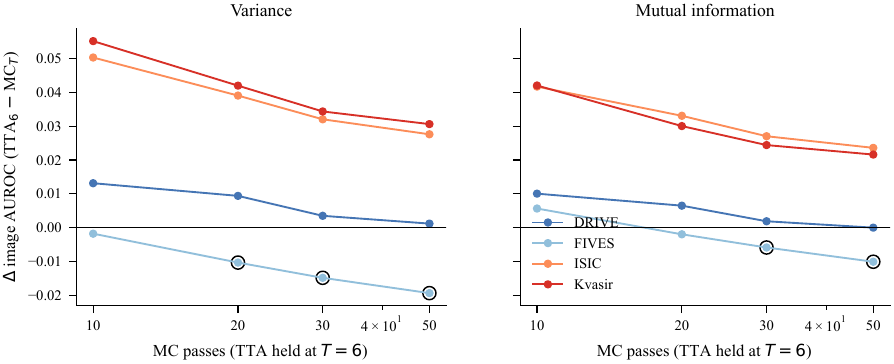}
\caption{\textbf{Unequal compute changes the MC--TTA comparison.} TTA is held
at six passes while MC Dropout is increased to 10, 20, 30, and 50 stochastic
passes. DRIVE approaches a tie, FIVES crosses to MC, and ISIC/Kvasir remain
TTA-favourable in the tested settings.}
\label{fig:supp-budget-crossing}
\end{figure*}

Unequal-budget comparisons are intentionally adverse to a universal TTA claim.
They show that estimator ranking can depend on the allowed inference budget. We
therefore state the MC--TTA result as a matched-budget finding rather than a
compute-independent ordering.

\subsection{Deep ensembles and estimator-family effects}
\label{supp:ensembles}

Deep ensembles provide a stronger estimator-family comparison. Unlike MC
Dropout and TTA, a deep ensemble combines independently trained models and
therefore changes the trained segmenter family. We report it separately from the
within-checkpoint MC--TTA comparison.

\begin{figure*}[t]
\centering
\includegraphics[width=0.78\textwidth]{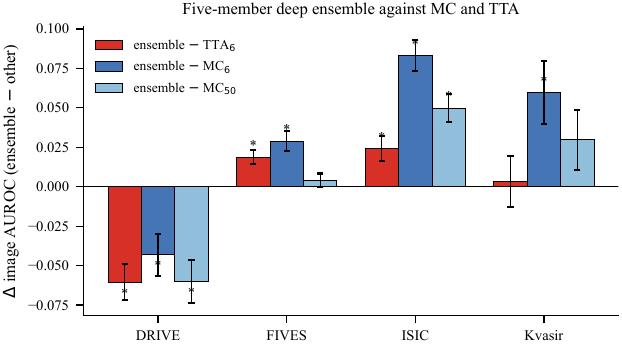}
\caption{\textbf{Estimator family can change the winner.} Five-member
deterministic deep ensembles outperform TTA on FIVES and ISIC, underperform
it on DRIVE, and are statistically indistinguishable from TTA on Kvasir after
correction.}
\label{fig:supp-estimator-family}
\end{figure*}

% Generated by wacv_r2_final/scripts/run_05_build_tables.py. Do not edit by hand.
\begin{table}[t]
\centering
\small
\setlength{\tabcolsep}{2.2pt}
\caption{\textbf{No estimator family wins everywhere once budget is unconstrained.}
Mean image AUROC, same images and backbone throughout; TTA uses 6 passes. $\Delta$
columns are paired differences \emph{against TTA-T6}, so positive means the competitor
is better; \textsuperscript{*} is $q<0.05$ within each column's family. At $8\times$
the passes MC ties TTA on DRIVE and overtakes it on FIVES; deep ensembles beat TTA on
FIVES and ISIC and lose on DRIVE.}
\label{tab:estimator-family}
\resizebox{\linewidth}{!}{%
\begin{tabular}{llcccccc}
\toprule
 & & \multicolumn{4}{c}{mean image AUROC} & \multicolumn{2}{c}{$\Delta$ vs TTA-T6} \\
\cmidrule(lr){3-6}\cmidrule(lr){7-8}
Dataset & Func. & MC T6 & MC T50 & TTA T6 & Ens.\ 5 & MC T50 & Ens.\ 5 \\
\midrule
DRIVE & Var & 0.839 & 0.857 & 0.858 & 0.806 & -0.001 & -0.052\textsuperscript{*} \\
DRIVE & MI & 0.852 & 0.868 & 0.868 & 0.799 & +0.000 & -0.069\textsuperscript{*} \\
FIVES & Var & 0.852 & 0.878 & 0.858 & 0.881 & +0.019\textsuperscript{*} & +0.023\textsuperscript{*} \\
FIVES & MI & 0.859 & 0.882 & 0.872 & 0.886 & +0.010\textsuperscript{*} & +0.014\textsuperscript{*} \\
ISIC & Var & 0.803 & 0.840 & 0.867 & 0.896 & -0.028\textsuperscript{*} & +0.029\textsuperscript{*} \\
ISIC & MI & 0.821 & 0.851 & 0.874 & 0.894 & -0.024\textsuperscript{*} & +0.019\textsuperscript{*} \\
Kvasir & Var & 0.832 & 0.866 & 0.897 & 0.908 & -0.031\textsuperscript{*} & +0.011 \\
Kvasir & MI & 0.864 & 0.889 & 0.911 & 0.907 & -0.022\textsuperscript{*} & -0.004 \\
\bottomrule
\end{tabular}}
\end{table}

The ensemble results rule out a universal estimator ordering. Ensembles are
significantly better than TTA on FIVES and ISIC, significantly worse on DRIVE,
and not significantly different on Kvasir after correction. This supports the
main paper's claim that uncertainty should be evaluated as an
estimator--policy--review configuration rather than by assuming one estimator
family is always preferred.

\subsection{Learned deferral and policy transfer}
\label{supp:learned-transfer}

Learned deferral heads estimate error probability from validation errors while
keeping the segmentation backbone frozen. The deployable protocol fits the head
and any adaptive thresholds on the cell's validation split and evaluates on the
held-out test set. Test labels are never used to fit the deployable policy.

\begin{table}[t]
\centering
\small
\setlength{\tabcolsep}{4pt}
\caption{\textbf{Learned-deferral accounting.} The deployable head is fitted on the held-out validation split and evaluated on test. The separate five-fold test-distribution cross-fit is retained only as a diagnostic upper bound.}
\label{tab:supp-learned-accounting}
\begin{tabular}{lrrrr}
\toprule
Dataset & Train & Val / fit imgs & Test & Capt.@20 wins \\
\midrule
DRIVE & 16 & 4 & 20 & 2/2 \\
FIVES & 520 & 120 & 160 & 2/4 \\
ISIC & 1686 & 389 & 519 & 4/4 \\
Kvasir & 650 & 150 & 200 & 8/8 \\
\midrule
Total & -- & -- & -- & 16/18 \\
\bottomrule
\end{tabular}
\end{table}

Validation size is an important part of the evidence. DRIVE has only four
validation images for the deployable learned head. Although this provides many
pixels, it is image-starved and cannot demonstrate coverage of anatomical
variation. For this reason, learned-deferral results are reported together with
their fitting data.

\begin{table*}[t]
\centering
\scriptsize
\setlength{\tabcolsep}{3pt}
\caption{\textbf{Deployable validation-fitted learned deferral at Capt.@20\%.} The head is fitted only on each cell's validation split and evaluated on test. $\Delta=$ learned minus plain uncertainty; $^{*}$ denotes BH $q<.05$ within the corresponding estimator family.}
\label{tab:supp-learned-capture20-deployable}
\begin{tabular}{llrrccccr}
\toprule
Cell & Arm & Fit imgs & Eval imgs & Learned & Plain $u$ & $\Delta$ & 95\% CI & $q$ \\
\midrule
drive/unet & MC & 4 & 20 & 0.796 & 0.623 & +0.173$ ^{*}$ & [+0.133,+0.218] & $<.001$ \\
drive/unet & TTA & 4 & 20 & 0.830 & 0.817 & +0.013$ ^{*}$ & [+0.010,+0.015] & $<.001$ \\
fives/unet & MC & 120 & 160 & 0.861 & 0.790 & +0.071$ ^{*}$ & [+0.059,+0.084] & $<.001$ \\
fives/unet & TTA & 120 & 160 & 0.924 & 0.928 & -0.004$ ^{*}$ & [-0.004,-0.003] & $<.001$ \\
fives/unetpp & MC & 120 & 160 & 0.838 & 0.778 & +0.061$ ^{*}$ & [+0.053,+0.069] & $<.001$ \\
fives/unetpp & TTA & 120 & 160 & 0.908 & 0.909 & -0.001$ $ & [-0.002,+0.001] & 0.337 \\
isic/deeplabv3plus & MC & 389 & 519 & 0.828 & 0.814 & +0.014$ ^{*}$ & [+0.013,+0.016] & $<.001$ \\
isic/deeplabv3plus & TTA & 389 & 519 & 0.921 & 0.905 & +0.016$ ^{*}$ & [+0.013,+0.020] & $<.001$ \\
isic/unet & MC & 389 & 519 & 0.901 & 0.772 & +0.129$ ^{*}$ & [+0.119,+0.140] & $<.001$ \\
isic/unet & TTA & 389 & 519 & 0.936 & 0.932 & +0.004$ ^{*}$ & [+0.002,+0.008] & 0.003 \\
kvasir/deeplabv3plus & MC & 150 & 200 & 0.808 & 0.767 & +0.041$ ^{*}$ & [+0.035,+0.046] & $<.001$ \\
kvasir/deeplabv3plus & TTA & 150 & 200 & 0.911 & 0.907 & +0.004$ ^{*}$ & [+0.001,+0.007] & 0.001 \\
kvasir/transunet & MC & 150 & 200 & 0.781 & 0.646 & +0.136$ ^{*}$ & [+0.118,+0.154] & $<.001$ \\
kvasir/transunet & TTA & 150 & 200 & 0.881 & 0.782 & +0.099$ ^{*}$ & [+0.089,+0.109] & $<.001$ \\
kvasir/unet & MC & 150 & 200 & 0.868 & 0.580 & +0.288$ ^{*}$ & [+0.250,+0.326] & $<.001$ \\
kvasir/unet & TTA & 150 & 200 & 0.939 & 0.925 & +0.013$ ^{*}$ & [+0.008,+0.020] & $<.001$ \\
kvasir/unetpp & MC & 150 & 200 & 0.760 & 0.654 & +0.105$ ^{*}$ & [+0.082,+0.129] & $<.001$ \\
kvasir/unetpp & TTA & 150 & 200 & 0.906 & 0.893 & +0.013$ ^{*}$ & [+0.010,+0.016] & $<.001$ \\
\bottomrule
\end{tabular}
\end{table*}

Under the deployable protocol, the learned head improves Capt.@20\% in 16/18
cell--estimator settings, is significantly worse than plain uncertainty in one,
and is statistically tied in one. The gains are largest for weaker MC-based
rankings. On AURC, however, plain uncertainty remains ahead in some settings, so
a learned policy is not a uniformly dominant replacement for raw uncertainty.

Policy transfer is evaluated by fitting source policies only on source
validation data and measuring performance on held-out target test images, with
zero source-validation/target-test overlap. We compare transfer with two target
references: raw target uncertainty and an in-domain learned refit.

\begin{table*}[t]
\centering
\scriptsize
\setlength{\tabcolsep}{2.4pt}
\caption{\textbf{Clean TTA policy transfer versus raw target uncertainty (AURC, 18 pairs). Source policies are fitted on source validation data and evaluated on the full target test set. AURC is lower-is-better, so negative $\Delta=$ transfer minus reference favours transfer.} $^{*}$ denotes BH $q<.05$ across the 18-pair family.}
\label{tab:supp-transfer-raw}
\resizebox{\textwidth}{!}{%
\begin{tabular}{lllrrrrrrl}
\toprule
Arm & Source & Target & $n$ & Transfer & Reference & $\Delta$ & 95\% CI & $q$ & Verdict \\
\midrule
A cross domain & kvasir/unetpp & fives/unetpp & 160 & 0.0023 & 0.0022 & +0.0002$ ^{*}$ & [+0.0001,+0.0002] & $<.001$ & Reference \\
A cross domain & fives/unetpp & kvasir/unetpp & 200 & 0.0236 & 0.0082 & +0.0154$ ^{*}$ & [+0.0102,+0.0212] & $<.001$ & Reference \\
A cross domain & kvasir/deeplabv3plus & isic/deeplabv3plus & 519 & 0.0074 & 0.0081 & -0.0008$ ^{*}$ & [-0.0011,-0.0004] & $<.001$ & Transfer \\
A cross domain & isic/deeplabv3plus & kvasir/deeplabv3plus & 200 & 0.0083 & 0.0084 & -0.0001$ $ & [-0.0002,+0.0000] & 0.086 & n.s. \\
A cross domain & drive/unet & fives/unet & 160 & 0.0019 & 0.0018 & +0.0001$ ^{*}$ & [+0.0000,+0.0001] & $<.001$ & Reference \\
A cross domain & fives/unet & drive/unet & 20 & 0.0173 & 0.0151 & +0.0021$ ^{*}$ & [+0.0019,+0.0024] & $<.001$ & Reference \\
A cross domain & isic/unet & kvasir/unet & 200 & 0.0043 & 0.0043 & -0.0001$ $ & [-0.0003,+0.0001] & 0.552 & n.s. \\
A cross domain & kvasir/unet & isic/unet & 519 & 0.0042 & 0.0045 & -0.0003$ ^{*}$ & [-0.0005,-0.0001] & $<.001$ & Transfer \\
B cross backbone & kvasir/unet & kvasir/unetpp & 200 & 0.0078 & 0.0082 & -0.0004$ ^{*}$ & [-0.0008,-0.0001] & 0.003 & Transfer \\
B cross backbone & kvasir/unetpp & kvasir/unet & 200 & 0.0044 & 0.0043 & +0.0001$ $ & [-0.0001,+0.0003] & 0.634 & n.s. \\
B cross backbone & kvasir/unet & kvasir/deeplabv3plus & 200 & 0.0083 & 0.0084 & -0.0001$ ^{*}$ & [-0.0002,-0.0000] & 0.028 & Transfer \\
B cross backbone & kvasir/deeplabv3plus & kvasir/unet & 200 & 0.0042 & 0.0043 & -0.0002$ ^{*}$ & [-0.0003,-0.0000] & 0.013 & Transfer \\
B cross backbone & kvasir/unetpp & kvasir/deeplabv3plus & 200 & 0.0083 & 0.0084 & -0.0001$ ^{*}$ & [-0.0002,-0.0000] & 0.026 & Transfer \\
B cross backbone & kvasir/deeplabv3plus & kvasir/unetpp & 200 & 0.0080 & 0.0082 & -0.0002$ ^{*}$ & [-0.0003,-0.0001] & $<.001$ & Transfer \\
B cross backbone & fives/unet & fives/unetpp & 160 & 0.0023 & 0.0022 & +0.0001$ ^{*}$ & [+0.0001,+0.0002] & $<.001$ & Reference \\
B cross backbone & fives/unetpp & fives/unet & 160 & 0.0018 & 0.0018 & +0.0001$ ^{*}$ & [+0.0000,+0.0001] & $<.001$ & Reference \\
B cross backbone & isic/unet & isic/deeplabv3plus & 519 & 0.0067 & 0.0081 & -0.0014$ ^{*}$ & [-0.0020,-0.0008] & $<.001$ & Transfer \\
B cross backbone & isic/deeplabv3plus & isic/unet & 519 & 0.0067 & 0.0045 & +0.0021$ ^{*}$ & [+0.0010,+0.0033] & $<.001$ & Reference \\
\bottomrule
\end{tabular}}
\end{table*}

\begin{table*}[t]
\centering
\scriptsize
\setlength{\tabcolsep}{2.4pt}
\caption{\textbf{Clean TTA policy transfer versus an in-domain refit (AURC, 18 pairs). Negative $\Delta=$ transfer minus reference favours transfer. The mixed result shows that target-domain refitting is usually safer than transferring the source policy.} $^{*}$ denotes BH $q<.05$ across the 18-pair family.}
\label{tab:supp-transfer-refit}
\resizebox{\textwidth}{!}{%
\begin{tabular}{lllrrrrrrl}
\toprule
Arm & Source & Target & $n$ & Transfer & Reference & $\Delta$ & 95\% CI & $q$ & Verdict \\
\midrule
A cross domain & kvasir/unetpp & fives/unetpp & 160 & 0.0023 & 0.0021 & +0.0002$ ^{*}$ & [+0.0002,+0.0003] & $<.001$ & Reference \\
A cross domain & fives/unetpp & kvasir/unetpp & 200 & 0.0236 & 0.0082 & +0.0153$ ^{*}$ & [+0.0102,+0.0211] & $<.001$ & Reference \\
A cross domain & kvasir/deeplabv3plus & isic/deeplabv3plus & 519 & 0.0074 & 0.0059 & +0.0015$ ^{*}$ & [+0.0007,+0.0023] & $<.001$ & Reference \\
A cross domain & isic/deeplabv3plus & kvasir/deeplabv3plus & 200 & 0.0083 & 0.0082 & +0.0001$ $ & [-0.0000,+0.0002] & 0.100 & n.s. \\
A cross domain & drive/unet & fives/unet & 160 & 0.0019 & 0.0020 & -0.0001$ ^{*}$ & [-0.0002,-0.0001] & $<.001$ & Transfer \\
A cross domain & fives/unet & drive/unet & 20 & 0.0173 & 0.0126 & +0.0047$ ^{*}$ & [+0.0039,+0.0056] & $<.001$ & Reference \\
A cross domain & isic/unet & kvasir/unet & 200 & 0.0043 & 0.0042 & +0.0001$ $ & [-0.0001,+0.0003] & 0.507 & n.s. \\
A cross domain & kvasir/unet & isic/unet & 519 & 0.0042 & 0.0052 & -0.0010$ $ & [-0.0020,-0.0000] & 0.054 & n.s. \\
B cross backbone & kvasir/unet & kvasir/unetpp & 200 & 0.0078 & 0.0082 & -0.0005$ ^{*}$ & [-0.0010,-0.0001] & 0.049 & Transfer \\
B cross backbone & kvasir/unetpp & kvasir/unet & 200 & 0.0044 & 0.0042 & +0.0002$ $ & [-0.0000,+0.0005] & 0.106 & n.s. \\
B cross backbone & kvasir/unet & kvasir/deeplabv3plus & 200 & 0.0083 & 0.0082 & +0.0001$ ^{*}$ & [+0.0000,+0.0001] & $<.001$ & Reference \\
B cross backbone & kvasir/deeplabv3plus & kvasir/unet & 200 & 0.0042 & 0.0042 & -0.0000$ $ & [-0.0002,+0.0001] & 0.852 & n.s. \\
B cross backbone & kvasir/unetpp & kvasir/deeplabv3plus & 200 & 0.0083 & 0.0082 & +0.0001$ ^{*}$ & [+0.0000,+0.0002] & 0.004 & Reference \\
B cross backbone & kvasir/deeplabv3plus & kvasir/unetpp & 200 & 0.0080 & 0.0082 & -0.0002$ ^{*}$ & [-0.0004,-0.0001] & 0.009 & Transfer \\
B cross backbone & fives/unet & fives/unetpp & 160 & 0.0023 & 0.0021 & +0.0002$ ^{*}$ & [+0.0001,+0.0002] & $<.001$ & Reference \\
B cross backbone & fives/unetpp & fives/unet & 160 & 0.0018 & 0.0020 & -0.0002$ ^{*}$ & [-0.0002,-0.0002] & $<.001$ & Transfer \\
B cross backbone & isic/unet & isic/deeplabv3plus & 519 & 0.0067 & 0.0059 & +0.0008$ ^{*}$ & [+0.0003,+0.0014] & $<.001$ & Reference \\
B cross backbone & isic/deeplabv3plus & isic/unet & 519 & 0.0067 & 0.0052 & +0.0015$ ^{*}$ & [+0.0012,+0.0018] & $<.001$ & Reference \\
\bottomrule
\end{tabular}}
\end{table*}

Across 18 source--target pairs, transferred TTA policies significantly beat raw
target uncertainty in 8 pairs, significantly lose in 7, and are not significant
in 3. Against an in-domain refit, transfer is significantly better in 4 pairs,
significantly worse in 9, and not significant in 5. Transfer is therefore
partial: it can help between related cells, but target-domain uncertainty and
in-domain fitting remain strong baselines.

\subsection{Review-assumption sensitivity}
\label{supp:review-assumptions}

Fixed-budget deferral assumes that deferred units can be corrected. This is an
analytical review proxy rather than a reader-study measurement. We therefore
audit whether estimator rankings remain stable when the review assumption
changes while the underlying prediction artifacts are held fixed.

\begin{figure}[t]
\centering
\includegraphics[width=\linewidth]{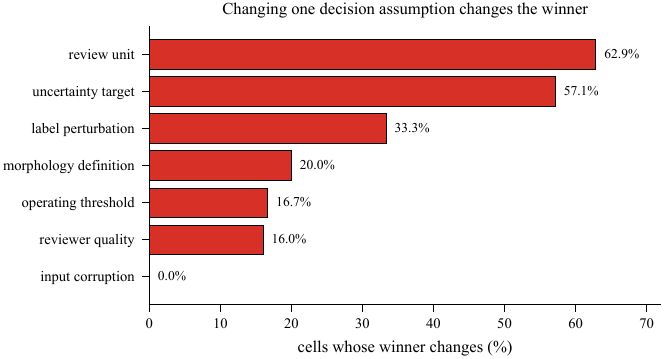}
\caption{\textbf{Decision-layer reversal rates on the DRIVE anchor.}
Predictions are held fixed while one review assumption is varied. These are
finite-grid sensitivity summaries, not population rates.}
\label{fig:supp-decision}
\end{figure}

\begin{table*}[t]
\centering
\scriptsize
\setlength{\tabcolsep}{2pt}
\caption{\textbf{CPU decision-assumption campaign.} All analyses reuse stored prediction artifacts. \texttt{PASS\_WITH\_LIMITATION} is retained rather than collapsed into PASS; the separately planned cross-domain EXP14 is blocked because comparable non-DRIVE validation maps were not retained and is not treated as a null result.}
\label{tab:supp-cpu-campaign}
\resizebox{\textwidth}{!}{%
\begin{tabular}{llllllll}
\toprule
ID & Experiment & Status & Data & Method & Metric & Effect & CI \\
\midrule
CPU-001 & exp01\_drive\_annotation\_disagreement & PASS & DRIVE & mc\_mutual\_information & mean\_image\_auroc\_human\_disagreement & 0.8131 & [0.8007, 0.8253] \\
CPU-002 & exp01\_drive\_annotation\_disagreement & PASS & DRIVE & tta\_mutual\_information & mean\_image\_auroc\_human\_disagreement & 0.8166 & [0.8056, 0.8271] \\
CPU-003 & exp02\_imperfect\_reviewer & PASS & DRIVE & all\_estimators & ranking\_reversal\_fraction\_vs\_perfect\_reviewer & 0.1600 & -- \\
CPU-004 & exp03\_review\_unit\_sensitivity & PASS & DRIVE & all\_estimators & fraction\_review\_unit\_budget\_cells\_reversing\_pixel\_winner & 0.6286 & -- \\
CPU-005 & exp04\_calibration & PASS\_WITH\_LIMITATION & DRIVE & mc & best\_calibrated\_probability\_minus\_best\_sampling\_error\_auroc & -0.0506 & -- \\
CPU-006 & exp04\_calibration & PASS\_WITH\_LIMITATION & DRIVE & tta & best\_calibrated\_probability\_minus\_best\_sampling\_error\_auroc & -0.0315 & -- \\
CPU-007 & exp05\_crossfit\_deferral & PASS\_WITH\_LIMITATION & DRIVE & tta\_ridge & learned\_minus\_plain\_capture\_at\_20\% & 0.0308 & -- \\
CPU-008 & exp06\_drive\_corruption\_robustness & PASS & DRIVE & mc\_t6\_vs\_tta\_t6 & fraction\_corruption\_score\_cells\_reversing\_clean\_winner & 0.0000 & -- \\
CPU-009 & exp07\_uncertainty\_target\_sensitivity & PASS & DRIVE & all\_estimators & fraction\_targets\_reversing\_all\_error\_winner & 0.5714 & -- \\
CPU-010 & exp08\_cpu\_reproducibility & PASS\_WITH\_LIMITATION & DRIVE & tta\_t6 & cpu\_minus\_gpu\_mean\_image\_error\_auroc & -0.0003 & [-0.0008, 0.0003] \\
CPU-011 & exp09\_threshold\_sensitivity & PASS & DRIVE & all\_estimators & fraction\_threshold\_settings\_reversing\_canonical\_winner & 0.1667 & -- \\
CPU-012 & exp10\_label\_perturbation\_robustness & PASS\_WITH\_LIMITATION & DRIVE & all\_estimators & fraction\_label\_targets\_reversing\_canonical\_winner & 0.3333 & -- \\
CPU-013 & exp16\_spatial\_morphology\_sensitivity & PASS & DRIVE & all\_estimators & fraction\_region\_definitions\_reversing\_canonical\_winner & 0.2000 & -- \\
CPU-014 & exp11\_component\_level\_decisions & PASS\_WITH\_LIMITATION & KVASIR & mc\_variance & image\_level\_auroc\_spurious\_pred\_component & 0.9784 & -- \\
CPU-015 & exp12\_cost\_utility\_sweep & PASS & DRIVE & tta\_probability\_baseline & method\_preference\_fraction\_across\_cost\_grid & 0.6333 & -- \\
CPU-016 & exp13\_calibration\_under\_shift & PASS\_WITH\_LIMITATION & DRIVE & deterministic\_clean\_fitted\_calibrators & fraction\_shifted\_conditions\_improving\_ece\_vs\_uncalibrated & 0.0357 & -- \\
CPU-017 & exp15\_classical\_probability\_uncertainty & PASS & DRIVE & best\_sampling\_vs\_best\_probability & mean\_image\_error\_auroc\_difference & 0.0220 & -- \\
CPU-018 & exp17\_referral\_normalization\_stability & PASS & DRIVE & all\_estimators & maximum\_absolute\_capture\_change\_vs\_raw\_score\_allocation & 0.1883 & -- \\
\bottomrule
\end{tabular}}
\end{table*}

Changing the review unit and the uncertainty target produces the largest ranking
changes. Reviewer quality, threshold choice, label perturbation, morphology, and
cost/utility assumptions also change preferred estimators in some settings,
whereas the tested corruption grid does not reverse the MC--TTA ordering. These
checks show that the review model is not a secondary implementation detail; it
is part of the evaluated decision configuration.

\subsection{Boundary, thin-structure, and second-observer analysis}
\label{supp:observer}

Structural stratification separates global error detection into boundary,
interior, and thin-structure regions. Boundaries are defined by a morphological
band around the foreground boundary, and thin structures are defined by
erosion-based foreground thinning. Because no single morphology radius is
canonical, we report radius sensitivity rather than treating one radius as a
universal definition.

\begin{figure*}[t]
\centering
\includegraphics[width=0.95\textwidth]{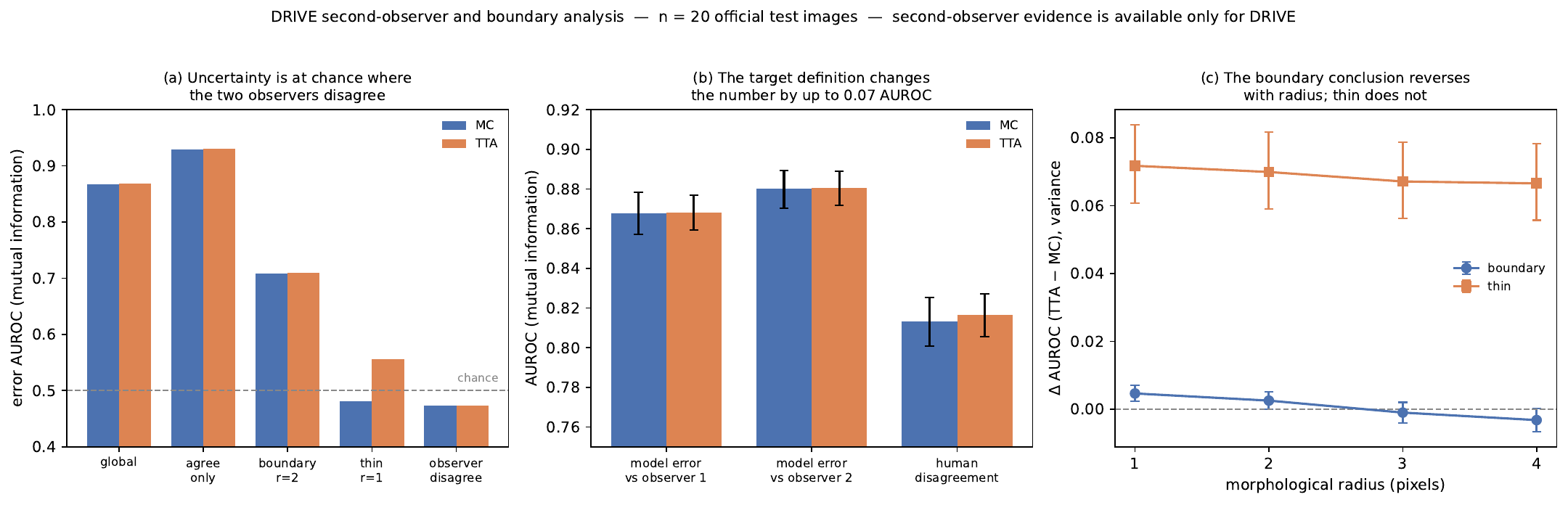}
\caption{\textbf{DRIVE second-observer audit.} Disagreement is the XOR of the
two manual annotations inside the DRIVE field of view. Human disagreement is
concentrated near vessel boundaries, while uncertainty is near chance on the
disputed pixels themselves.}
\label{fig:supp-observer}
\end{figure*}

\begin{table*}[t]
\centering
\small
\setlength{\tabcolsep}{4pt}
\caption{\textbf{DRIVE second-observer MC--TTA audit ($n=20$).} Scores detect model error with respect to observer 2. $\Delta=$TTA$-$MC. Global MC/TTA differences are negligible; the robust positive difference is concentrated on thin structures. Second-observer evidence is DRIVE-only.}
\label{tab:supp-observer2-detailed}
\begin{tabular}{llccccc}
\toprule
Score & Region & MC & TTA & $\Delta$ & 95\% CI & $q$ \\
\midrule
variance & Global & 0.857 & 0.859 & +0.001$ $ & [-0.005,+0.009] & 0.862 \\
variance & DisAgree & 0.474 & 0.473 & -0.001$ $ & [-0.008,+0.006] & 0.971 \\
variance & Agree & 0.922 & 0.924 & +0.002$ $ & [-0.001,+0.006] & 0.444 \\
variance & Boundary $r=2$ & 0.714 & 0.717 & +0.003$ $ & [-0.000,+0.005] & 0.164 \\
variance & Thin $r=1$ & 0.514 & 0.585 & +0.072$ ^{*}$ & [+0.061,+0.084] & $<.001$ \\
MI & Global & 0.868 & 0.868 & +0.000$ $ & [-0.005,+0.007] & 0.971 \\
MI & DisAgree & 0.473 & 0.473 & +0.000$ $ & [-0.007,+0.007] & 0.994 \\
MI & Agree & 0.929 & 0.931 & +0.001$ $ & [-0.002,+0.005] & 0.599 \\
MI & Boundary $r=2$ & 0.708 & 0.710 & +0.002$ $ & [-0.001,+0.005] & 0.435 \\
MI & Thin $r=1$ & 0.481 & 0.555 & +0.075$ ^{*}$ & [+0.064,+0.087] & $<.001$ \\
Prob.-derived & Global & 0.817 & 0.823 & +0.006$ $ & [+0.001,+0.011] & 0.063 \\
Prob.-derived & DisAgree & 0.475 & 0.473 & -0.002$ $ & [-0.006,+0.003] & 0.599 \\
Prob.-derived & Agree & 0.892 & 0.901 & +0.009$ ^{*}$ & [+0.005,+0.013] & $<.001$ \\
Prob.-derived & Boundary $r=2$ & 0.721 & 0.732 & +0.011$ ^{*}$ & [+0.009,+0.013] & $<.001$ \\
Prob.-derived & Thin $r=1$ & 0.578 & 0.621 & +0.043$ ^{*}$ & [+0.035,+0.050] & $<.001$ \\
\bottomrule
\end{tabular}
\end{table*}

\begin{table*}[t]
\centering
\scriptsize
\setlength{\tabcolsep}{2.5pt}
\caption{\textbf{Boundary/thin-region radius sweep (variance AUROC).} $\Delta=$TTA$-$MC. The DRIVE boundary sign changes at radii 3--4, whereas the thin-structure advantage remains positive. Other datasets are included to expose radius sensitivity rather than canonize one radius.}
\label{tab:supp-radius-full}
\begin{tabular}{llrrrrrrr}
\toprule
Dataset & Region & $r$ & $n$ & Region frac. & MC & TTA & $\Delta$ & 95\% CI \\
\midrule
DRIVE & Boundary & 1 & 20 & 0.154 & 0.621 & 0.625 & +0.005 & [+0.002,+0.007] \\
DRIVE & Boundary & 2 & 20 & 0.267 & 0.714 & 0.717 & +0.003 & [-0.000,+0.005] \\
DRIVE & Boundary & 3 & 20 & 0.356 & 0.761 & 0.760 & -0.001 & [-0.004,+0.002] \\
DRIVE & Boundary & 4 & 20 & 0.429 & 0.786 & 0.783 & -0.003 & [-0.007,+0.000] \\
DRIVE & Thin & 1 & 20 & 0.071 & 0.514 & 0.585 & +0.072 & [+0.061,+0.084] \\
DRIVE & Thin & 2 & 20 & 0.107 & 0.624 & 0.694 & +0.070 & [+0.059,+0.082] \\
DRIVE & Thin & 3 & 20 & 0.122 & 0.668 & 0.735 & +0.067 & [+0.056,+0.079] \\
DRIVE & Thin & 4 & 20 & 0.126 & 0.678 & 0.745 & +0.067 & [+0.056,+0.078] \\
KVASIR & Boundary & 1 & 200 & 0.007 & 0.552 & 0.566 & +0.015 & [+0.009,+0.021] \\
KVASIR & Boundary & 2 & 200 & 0.014 & 0.595 & 0.618 & +0.022 & [+0.014,+0.030] \\
KVASIR & Boundary & 3 & 200 & 0.020 & 0.635 & 0.660 & +0.025 & [+0.016,+0.035] \\
KVASIR & Boundary & 4 & 200 & 0.027 & 0.670 & 0.696 & +0.026 & [+0.016,+0.036] \\
KVASIR & Thin & 1 & 200 & 0.004 & 0.422 & 0.473 & +0.054 & [+0.036,+0.071] \\
KVASIR & Thin & 2 & 200 & 0.007 & 0.491 & 0.547 & +0.059 & [+0.041,+0.076] \\
KVASIR & Thin & 3 & 200 & 0.010 & 0.562 & 0.613 & +0.053 & [+0.036,+0.070] \\
KVASIR & Thin & 4 & 200 & 0.014 & 0.623 & 0.667 & +0.045 & [+0.028,+0.063] \\
ISIC & Boundary & 1 & 519 & 0.010 & 0.550 & 0.557 & +0.007 & [+0.004,+0.010] \\
ISIC & Boundary & 2 & 519 & 0.018 & 0.587 & 0.602 & +0.015 & [+0.011,+0.019] \\
ISIC & Boundary & 3 & 519 & 0.026 & 0.616 & 0.635 & +0.018 & [+0.014,+0.023] \\
ISIC & Boundary & 4 & 519 & 0.034 & 0.640 & 0.661 & +0.021 & [+0.016,+0.026] \\
ISIC & Thin & 1 & 519 & 0.005 & 0.529 & 0.525 & +0.005 & [-0.009,+0.018] \\
ISIC & Thin & 2 & 519 & 0.009 & 0.549 & 0.552 & +0.010 & [-0.002,+0.021] \\
ISIC & Thin & 3 & 519 & 0.013 & 0.577 & 0.587 & +0.011 & [+0.000,+0.022] \\
ISIC & Thin & 4 & 519 & 0.017 & 0.606 & 0.619 & +0.013 & [+0.003,+0.023] \\
FIVES & Boundary & 1 & 160 & 0.077 & 0.731 & 0.773 & +0.043 & [+0.039,+0.046] \\
FIVES & Boundary & 2 & 160 & 0.137 & 0.823 & 0.859 & +0.036 & [+0.032,+0.040] \\
FIVES & Boundary & 3 & 160 & 0.184 & 0.859 & 0.886 & +0.027 & [+0.023,+0.031] \\
FIVES & Boundary & 4 & 160 & 0.223 & 0.874 & 0.894 & +0.020 & [+0.017,+0.024] \\
FIVES & Thin & 1 & 160 & 0.037 & 0.568 & 0.719 & +0.151 & [+0.141,+0.162] \\
FIVES & Thin & 2 & 160 & 0.059 & 0.664 & 0.818 & +0.153 & [+0.140,+0.166] \\
FIVES & Thin & 3 & 160 & 0.069 & 0.693 & 0.845 & +0.152 & [+0.139,+0.166] \\
FIVES & Thin & 4 & 160 & 0.072 & 0.698 & 0.852 & +0.154 & [+0.140,+0.168] \\
\bottomrule
\end{tabular}
\end{table*}

The two DRIVE observers have Dice 0.788 and disagree on 5.28\% of in-FOV pixels;
91.7\% of that disagreement lies inside the radius-2 boundary band. When the
target is human disagreement, mutual-information AUROC is 0.813 for MC and
0.817 for TTA. On the disputed pixels themselves, all evaluated uncertainty
scores are near chance. Thus, uncertainty can identify unstable regions, but it
does not adjudicate between two plausible annotations.

Annotation ambiguity does not fully explain the structural weakness. Under
observer-2 model-error labels, global MC--TTA differences are not significant,
while TTA's advantage is concentrated on thin structures. Boundary conclusions
change with the chosen radius, whereas the thin-region advantage remains
positive across the tested radii. The main paper therefore states the structural
claim narrowly: uncertainty is weakest near ambiguous boundaries and thin
structures, and this weakness is partly, but not entirely, explained by
annotation disagreement.

\subsection{Probability-derived baselines and calibration checks}
\label{supp:probability}

Probability-derived scores are included as mandatory cheap baselines. In binary
segmentation, entropy, inverse confidence, margin, distance from $1/2$, and
maximum-softmax confidence are monotone functions of $|\hat p-0.5|$ and induce
the same ranking up to reversal depending on whether the score is written as
confidence or uncertainty. They therefore produce the same fixed-budget deferred
set under a coverage-based policy.

On DRIVE TTA outputs, the maximum numerical identity deviation among these
binary scores is at floating-point scale, and the minimum Spearman correlation
among the resulting rankings is 1.000. Similarly, strictly monotone
transformations of a review score preserve the deferred set, risk--coverage
curve, and AURC. Calibration can improve the numerical interpretation of
probabilities without changing a coverage-based review decision when the ranking
is preserved. Calibration, threshold, label, and referral-normalisation analyses
are therefore interpreted as decision-layer sensitivity checks rather than as
independent estimator families.

\subsection{Qualitative examples}
\label{supp:qualitative}

\begin{figure*}[t]
\centering
\includegraphics[width=0.82\textwidth]{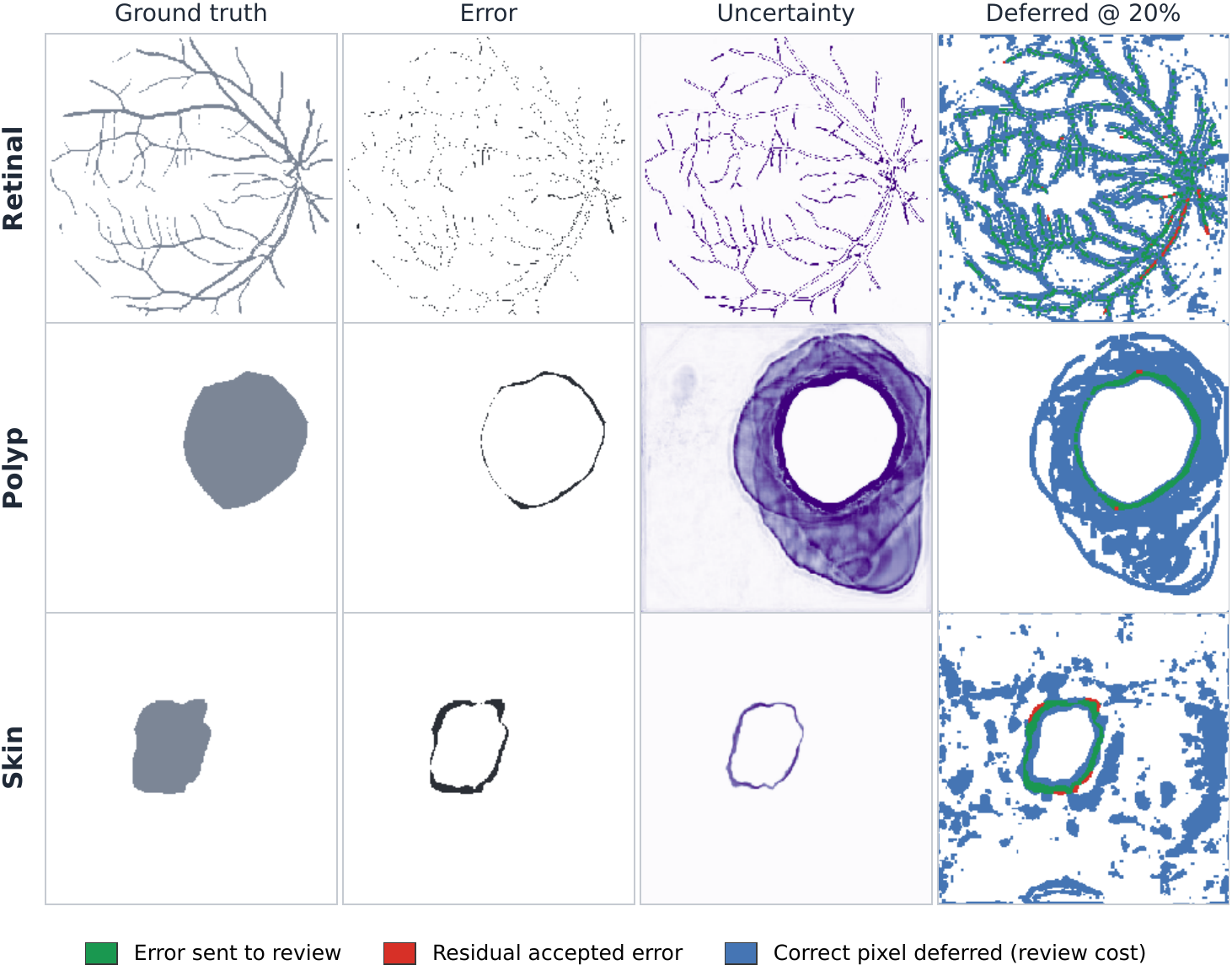}
\caption{\textbf{Representative qualitative examples across domains.}
Examples make the ranking problem concrete rather than supporting a new
quantitative claim. High uncertainty tends to concentrate around difficult
boundaries, small structures, and local segmentation failures, while different
estimators allocate review mass differently.}
\label{fig:supp-qualitative}
\end{figure*}

\Cref{fig:supp-qualitative} shows representative examples from the three segmentation domains. These examples are intended as qualitative checks on the decision rule rather than as additional evidence for a new quantitative claim. Across domains, uncertainty and deferral concentrate around the same structures identified by the quantitative analyses: vessel fragments and bifurcations in retinal images, object boundaries in polyp and lesion images, and local segmentation failures where the predicted mask disagrees with the annotation. The deferred set contains both true errors sent to review and correct pixels that consume review budget, illustrating why fixed-budget evaluation must report both error capture and accepted risk.

\subsection{Scope-limited, negative, and blocked evidence}
\label{supp:limits}

We retain negative and scope-limited findings rather than converting them into
positive claims. The main limitations are: no end-to-end reader study;
second-observer evidence only for DRIVE; dropout-rate and training-seed sweeps
only for DRIVE; self-defined FIVES, ISIC, and Kvasir splits; severe image
scarcity for deployable learned deferral on DRIVE; policy transfer that is
usually worse than an in-domain refit; deep-ensemble wins over TTA on FIVES and
ISIC; unequal-compute settings where MC can tie or exceed TTA; DRIVE boundary
conclusions that depend on morphology radius; and one blocked cross-domain
validation-map analysis due to missing comparable validation artifacts. None of
these findings is used to support a stronger claim than the evidence permits.

\subsection{Reproducibility package}
\label{supp:repro}

The anonymous release separates model generation from the lightweight decision
audit. It includes executed result archives, configuration records, split
metadata, per-image derived evidence, table and figure scripts, and a claim
ledger. The derived decision-audit pipeline can reproduce the reported tables
from saved artifacts without rerunning large model checkpoints. Artifact
manifests record image-ID pairing, thresholds, field-of-view policies, row
counts, ensemble-member counts, and table-generation inputs.

\end{document}